\documentclass[journal]{IEEEtran}

\ifCLASSINFOpdf

\else

\fi

\usepackage{graphicx}
\usepackage{booktabs}
\usepackage{tabularx}
\usepackage{float}
\usepackage{amsmath,amssymb} 
\usepackage{multirow}
\usepackage{color}
\usepackage[ruled,vlined]{algorithm2e}

\usepackage{blindtext}
\usepackage{chngcntr}

\newtheorem{theorem}{Theorem}

\begin{document}

\title{Automatic Analysis of Facial Expressions Based on Deep Covariance Trajectories}

\author{Naima~Otberdout,~\IEEEmembership{Member,~IEEE,}
        Anis~Kacem,~\IEEEmembership{Member,~IEEE,}
        Mohamed~Daoudi,~\IEEEmembership{Senior,~IEEE,}
        Lahoucine~Ballihi,~\IEEEmembership{Member,~IEEE,}
        and~Stefano~Berretti,~\IEEEmembership{Senior,~IEEE}
\thanks{N. Otberdout and L. Ballihi are with the LRIT - CNRST URAC 29, Mohammed V University in Rabat, Faculty of Sciences, Rabat, Morocco. e-mail: naima.otberdout@um5s.net.ma, lahoucine.ballihi@um5.ac.ma}
\thanks{A. Kacem is with SnT - Interdisciplinary Centre for Security, Reliability and Trust, University of Luxembourg. email: anis.kacem@uni.lu. This work has been done when he was a Ph.D student at IMT Lille-Douai} 
\thanks{M. Daoudi is with IMT Lille-Douai, University of Lille, CNRS, UMR 9189 CRIStAL, Lille, France. e-mail: mohamed.daoudi@imt-lille-douai.fr}
\thanks{S. Berretti is with the Department of Information Engineering, University of Florence, 
Florence, Italy. e-mail: stefano.berretti@unifi.it}
}


\maketitle

\begin{abstract}
In this paper, we propose a new approach for facial expression recognition using deep covariance descriptors. The solution is based on the idea of encoding local and global Deep Convolutional Neural Network (DCNN) features extracted from still images, in compact local and global covariance descriptors. The space geometry of the covariance matrices is that of Symmetric Positive Definite (SPD) matrices. By conducting the classification of static facial expressions using Support Vector Machine (SVM) with a valid Gaussian kernel on the SPD manifold, we show that deep covariance descriptors are more effective than the standard classification with fully connected layers and softmax. Besides, we propose a completely new and original solution to model the temporal dynamic of facial expressions as deep trajectories on the SPD manifold. As an extension of the classification pipeline of covariance descriptors, we apply SVM with valid positive definite kernels derived from global alignment for deep covariance trajectories classification. By performing extensive experiments on the Oulu-CASIA, CK+, SFEW and AFEW datasets, we show that both the proposed static and dynamic approaches achieve state-of-the-art performance for facial expression recognition outperforming many recent approaches. 
\end{abstract}


\IEEEpeerreviewmaketitle


\section{Introduction}
\label{sec:intro}
For a long time, automated Facial Expression Recognition (FER) has been studied in many computer vision researches. This is due to the vital role of facial expressions in social interaction, and the wide spectrum of their potential applications that go from human computer interaction to medical and psychological  investigations.
As in several other applications, hand-crafted features, including geometric descriptors (\emph{e.g.}, distances between landmarks) and appearance descriptors (\emph{e.g.}, LBP, SIFT, HOG, etc.), were designed for many years to find a powerful face representation allowing an efficient  analysis of facial expressions. Some works have also explored higher order relations such as the covariance descriptor to encode these low-level features. Recently, Deep Convolutional Neural Networks (DCNNs) have radically changed the way to address this problem and opened the door for a quite different approach. Instead of using hand-crafted features, DCNN models learn from large collections of data to automatically extract the relevant patterns for the problem at hand. 
\indent

One limitation of current DCNN models is due to the fully connected layers that flatten the features extracted from the convolution layers, thus completely losing the spatial relationships within the face. To tackle this problem, we propose to discard the fully connected layers after the training phase, and directly use the global and local features extracted from the convolution layers in different facial regions. 
The question is how to encode these features in a compact and discriminative representation for a more efficient classification than the one achieved globally by classical softmax. 
Motivated by the impressive performance of the covariance descriptors used as second-order representations in many computer vision tasks~\cite{tuzel:2006,tuzel2008pedestrian}, in this work we propose to encode local and global deep facial features in local and global covariance descriptors.
We demonstrate the benefits of this representation in facial expression recognition from static images or collections of static peak frames, and from video sequences. 
For static images, we represent each face with local and global covariance descriptors that reside on the Symmetric Positive Definite (SPD) manifold; then, we define a valid positive definite Gaussian kernel on this manifold to be used with an SVM for static facial expressions classification. Conducting a thorough set of experiments with different DCNN architectures, \emph{i.e.}, VGG-face~\cite{parkhi2015deep} and ExpNet~\cite{ding2017facenet2expnet}, we demonstrate that our approach outperforms classification with the classical softmax. 

\begin{figure*}[!th]
\label{fig:Overview_tnnls}
\centering
\includegraphics[width=0.68\linewidth]{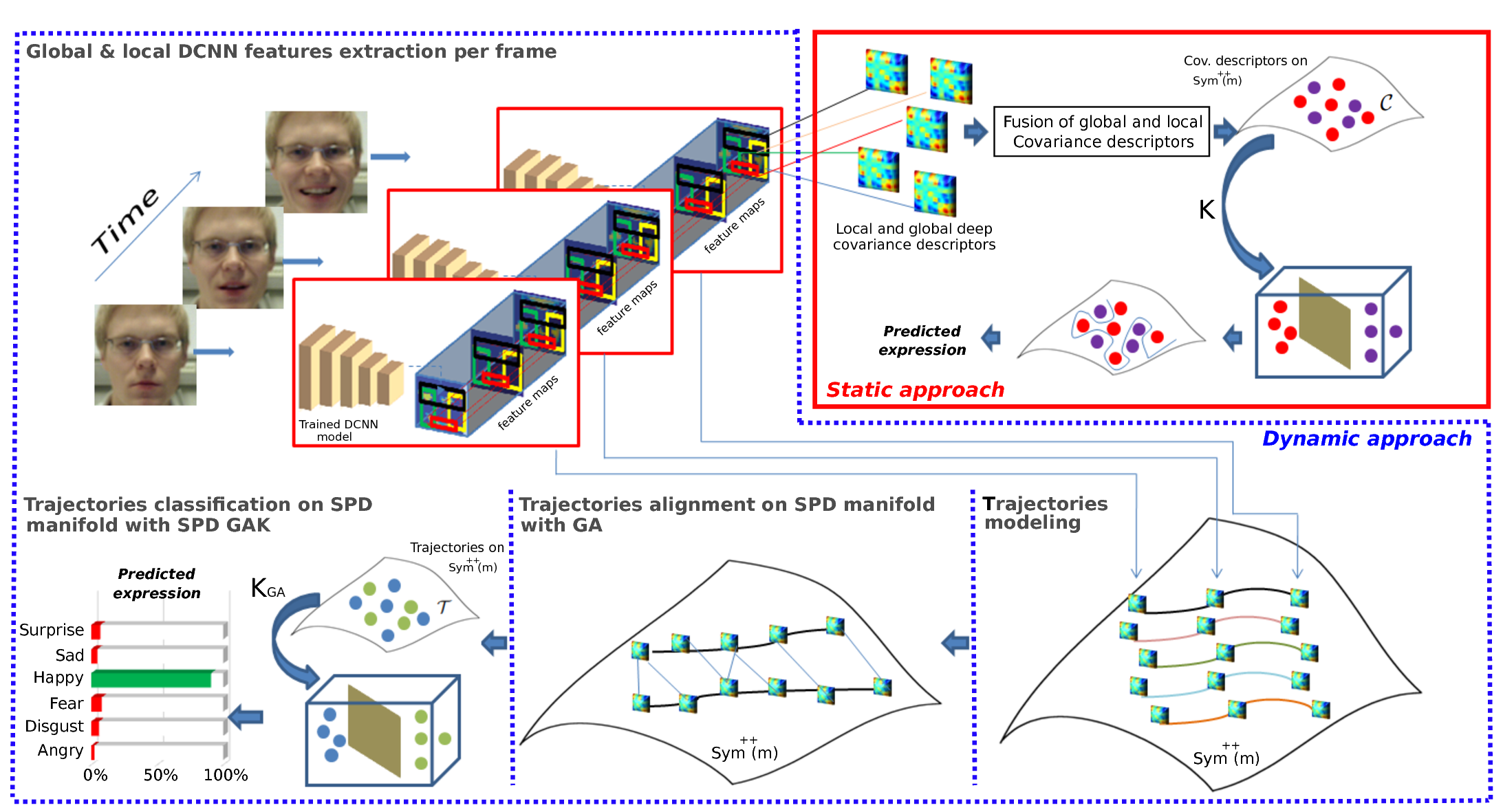}
\caption{Overview of the proposed approach. In the upper part, feature extraction and covariance matrix computation are illustrated on the left, while the static classification method on the SPD manifold is shown on the right. 
In the bottom part of the figure, the way trajectories are formed on the SPD manifold, and how they are aligned and classified is reported in the plots from right-to-left.}
\label{fig:overview-dynamic}
\end{figure*}

Furthermore, we extend our static approach to deal with dynamic facial expressions. The challenges encountered here are: how to represent the dynamic evolution of the video sequences? how to deal with the temporal misalignment of these videos to classify them in an efficient way? Regarding the first question, we exploit the space geometry of the covariance matrices as points on the SPD manifold, and model the temporal evolution of facial expressions as trajectories on this manifold. Following the static approach, we studied both global and local deep trajectories. Once constructing the deep trajectories, we need to align them in their manifold to remedy to the different execution rates of the facial expressions. A common method to do so is to use Dynamic Time Warping (DTW) as proposed in several works~\cite{amor2016action, kacem2018novel, gritai2009matching}. However, DTW does not define a proper metric and cannot be used to derive a valid positive-definite kernel for the classification phase~\cite{cuturi2007kernel}. 
Instead, in this work we propose global alignment of deep trajectories with the log-Euclidean Riemannian metric, which allows us to derive a valid positive-definite kernel used with SVM for the classification. By doing so, we propose a completely new approach to model and compare the spatial and the temporal evolution of facial expressions. 

 


Overall, our proposed method allows an efficient combination of both geometric and appearance features to define a compact representation of facial expressions, taking into consideration the spatial relationships within the face. 
In addition, this solution is extended to deal with both the spatial and the temporal domains of facial expressions. We illustrate in Figure~\ref{fig:Overview_tnnls}, an overview of the proposed approach.
In summary, the main contributions of this paper are: 
%
\begin{itemize}
\item Encoding local/global facial DCNN features by using local/global covariance matrices;
\item Using multiple late/early fusion schemes to combine multiple local and global information;
\item A temporal extension of the static covariance representations by modeling their temporal evolution as trajectories in the SPD manifold. To the best of our knowledge, this is the first work that uses DCNN features to model videos as trajectories on a Riemannian manifold; 
\item A temporal alignment method based on Global Alignment (GA), which is the first time to be proposed for aligning trajectories on the SPD manifold;
\item Classifying static facial expressions using a Gaussian kernel on the SPD manifold coupled with an SVM;
\item Classifying deep trajectories in the SPD manifold using a Global Alignment Kernel (GAK), which is a valid positive definite kernel, and an SVM;
\item Extensive experiments on three public datasets using two different DCNN architectures as well as a comparative study with the existing solutions.
\end{itemize}

We presented some preliminary ideas of this work in~\cite{Otberdout:2018}. With respect to~\cite{Otberdout:2018}, here we propose a completely new and original solution to model the temporal dynamic of facial expressions as trajectories on the SPD manifold. The experimental evaluation now comprises both the static and dynamic solutions, also including a larger number of datasets.

The remaining of the paper is organized as follows: In the next section, we present an overview of related works. In Section~\ref{sect:approach}, we introduce deep covariance descriptors as a way to encode deep facial features in a compact representation of the face; The way these descriptors can be used for expression classification from static images is reported in Section~\ref{sect:dcnn-classification}; In Section~\ref{sect:trajectories}, the approach is extended to the modeling of facial expressions as deep trajectories on the SPD manifold; In Section~\ref{sect:results}, we present an extensive experimentation of the proposed approaches as well as a comparison with the state-of-the-art; Lastly, conclusions and discussion are reported in Section~\ref{sect:conclusion}.

\section{Related work}\label{sect:related-work}
This section is organized into three parts; We first review works that use covariance descriptors for image/video classification; Then, we discuss works that employ DCNN features for static facial expression analysis, including some approaches that explore covariance descriptors to encode DCNN features; In the last part, we tackle the problem of facial expression recognition from dynamic data, by discussing solutions proposed for the temporal modeling of the facial expression evolution.

\textbf{Covariance Descriptors for Image/Video Classification:} 
In Computer Vision, covariance matrices have been shown to provide discriminative representations for both images and videos~\cite{tuzel:2006,ZhangSKLS18}. These solutions have shown impressive results in faces~\cite{pang2008gabor,harandi2012sparse} and actions~\cite{sanin2013spatio}, especially when accounting for the geometry of these representations as points in SPD manifold instead of handling them in the Euclidean space. 
However, using covariance matrices gives rise to many challenges and requires to develop effective and efficient inference methods. 
In~\cite{wang2012covariance}, Wang \emph{et al.} proposed Covariance Discriminative Learning (CDL) for image set classification. A covariance matrix is used to represent each image set and model the problem as classifying points on the Riemannian manifold spanned by non-singular covariance matrices. To take into account the geometry structure of covariance matrices, they derived a novel Riemannian kernel function, which successfully bridges the gap between traditional learning methods operating in vector spaces and the learning task on an unconventional manifold. In the same direction, Harandi \emph{et al.}~\cite{harandi2014manifold} proposed an approach to transform a high-dimensional SPD manifold into another SPD manifold with lower intrinsic dimension and maximum discriminative power. In~\cite{huang2015log}, Huang \emph{et al.} tackled the problem of the non-linear space by employing tangent space approximations. The method aims to learn a tangent map that can directly transform the matrix logarithms from the original tangent space to a new, more discriminant, tangent space. Their approach has been successfully applied to face recognition and face verification. 
Liu \emph{et al.}~\cite{liu2014combining} represented video clips by three types of image set models, \emph{i.e.}, linear subspace, covariance matrix, and Gaussian distribution, respectively, that can all be viewed as points residing on some Riemannian manifolds. Then, different Riemannian kernels were employed on these set models correspondingly for similarity/distance measurement. Kernel SVM, logistic regression, and partial least squares were investigated for classification. To further improve performance, an optimal fusion of classifiers is learned from different kernels and different modalities (video and audio) at the decision level.  

\textbf{DCNN for Static Facial Expression Recognition}: 
In the last few years, DCNN models have achieved a great success in different facial analysis tasks, including static facial expression recognition~\cite{mollahosseini2016going,ng2015deep}. The main challenge encountered when using DCNN models is the necessity of large-scale databases to train a good model. However, the databases available for facial expression recognition are quite small \emph{w.r.t.} other tasks. 
To address this challenge, some works opted for minimizing the depth and the complexity of the network by using a small deep architecture~\cite{khorrami2015deep}, while other works used deep models already trained on large expression datasets before fine-tuning them on the target dataset~\cite{ng2015deep,yu2015image}. To further boost the performance, Ding \emph{et al.}~\cite{ding2017facenet2expnet} proposed \textit{FaceNet2ExpNet}, which uses a very deep network trained for face recognition, to regularize a small deep network trained for facial expression recognition from static images. Many works opted for combining multiple DCNN models to further boost the results. 
For example, Kim \emph{et al.}~\cite{kim2015hierarchical} used a validation-accuracy-based exponentially-weighted average (VA-Expo-WA) rule to train multiple DCNN models by using different parameters of the models and adopting several learning strategies to use large external databases. In the same direction, Yu \emph{et al.}~\cite{yu2015image} combined two schemes for learning the ensemble weights of the network responses: by minimizing the log-likelihood loss, and by minimizing the hinge loss. 
However, all these works used a similar strategy, where a deep processing based on linear combinations, non-linearity activation and pooling are used to extract relevant features that are classified by fully connected and softmax layers. 
Taking a different direction, Yang \emph{et al.}~\cite{yang2018facial} proposed De-expression Residue Learning (DeRL), which consists of using Conditional Generative Adversarial Networks (cGANs) to filter out the expression of the person and provide its neutral image. By doing so, the expressive information is still encoded in the intermediate layers of cGAN and can be employed later on for expression classification. CGANs were also used by Yang \emph{et al.}~\cite{yang2018identity}. Given an identity, they proposed to generate six facial expressions given six trained cGAN networks. Then, the minimum distance between the input image and the generated expression images in the feature space was used to classify the expression of the input image. 
Besides, several other works introduced a novel class of DCNNs that explore second-order statistics (\emph{e.g.}, covariances). In the context of facial expression recognition from images, Acharya \emph{et al.}~\cite{acharya2018covariance} explored convolutional networks in conjunction with manifold networks for covariance pooling in an end-to-end deep learning manner. 
Wang \emph{et al.}~\cite{wang2017discriminative} presented Discriminative Covariance oriented Representation Learning (DCRL), which uses a DCNN model to project the face into a target feature space, while maximizing the discriminative ability of the covariance matrices calculated in this space. 

\textbf{Temporal Modeling of Facial Expressions}: 
The difficulty here is to account for the dynamic evolution of the facial expression. One direction to address this difficulty is to explore deep architectures that can model appearance and motion information simultaneously. For example, LSTMs combined with CNN have been successfully employed for facial expression recognition with different names such as CNN-RNN~\cite{fan2016video}, CNN-BRNN~\cite{yan2018multi}, etc. 
3D CNNs have also been used for facial expression recognition in several works including~\cite{fan2016video,liu2014deeply}. 
In the same direction, Jung \emph{et al.}~\cite{jung2015joint}, used a CNN to extract temporal appearance features from face image sequences with an additional deep network that extracts temporal geometry features from temporal facial landmarks. The two networks are then combined using a joint fine-tuning method. In~\cite{meng2016time}, Meng \emph{et al.} proposed Time-Delay Neural Network (TDNN) to model the temporal relationships between consecutive predictions on the decision level of a multistage system. This system was designed to continuously predict affective dimension values from facial expression videos. In~\cite{jan2018artificial}, Jan \emph{et al.} used different visual features including DCNN features to build a facial expression representation on the frame-level; then, feature dynamic history histogram (FDHH) was proposed to capture the temporal movement on the feature space. 
Acharya \emph{et al.}~\cite{acharya2018covariance} extended their static approach discussed before to dynamic facial expression recognition. They considered the temporal evolution of per-frame features by leveraging covariance pooling. Their networks achieve significant facial expression recognition performance for static data, while dynamic data are still more challenging.

Taking a different direction, several recent works chose to model the temporal evolution of the face as a trajectory. For example, Taheri \emph{et al.}~\cite{Taheri:2011} used landmark configurations of the face to represent facial deformations on the Grassmann manifold $G(2,n)$. They modeled the dynamics of facial expressions by parameterized trajectories on this manifold before classifying them using LDA followed by an SVM. In the same direction, Kacem \emph{et al.}~\cite{kacem:2017}, described the temporal evolution of facial landmarks as parameterized trajectories on the Riemannian manifold of positive semidefinite matrices of fixed-rank. Trajectories modeling in Riemannian manifolds was also used for human action recognition in several works~\cite{amor2016action,devanne:2015,chakraborty:2017}. However, all these works were based on geometric information to study the temporal evolution of some landmarks ignoring the texture information. 

One outstanding problem encountered when modeling the temporal evolution of the face as a trajectory is the temporal misalignment resulting from the different execution rate of the facial expression. This necessitates the use of an algorithm to align different trajectories, which is generally based on dynamic programming. Several works including~\cite{amor2016action,kacem2018novel,kacem:2017} used DTW to align trajectories in a Riemannian manifold; however, this algorithm does not define a proper metric, which is indeed required in the classification phase to define a valid positive-definite kernel. As alternative solution, different works~\cite{kacem2018novel,kacem:2017,gudmundsson2008support} proposed to ignore this constraint by using a variant of SVM with an arbitrary kernel without any restrictions on the kernel function. 

Different from the above methods, in this work, we use both global and local covariance descriptors computed on DCNN features to explore appearance and geometric features simultaneously. Furthermore, we propose a new solution for trajectories alignment in a Riemannian manifold based on global alignment. This allows us to derive a valid positive definite kernel for trajectory classification in the SPD manifold, instead of using an arbitrary kernel.

\section{Face representation}\label{sect:approach}
Given a set of $n_f$ face images $\mathcal{F} = \{f_{1}, f_{2},\dots,f_{n_f}\}$ labeled with their corresponding expressions $\{y_{1},y_{2},\dots,y_{n_f}\}$, we aim to find an efficient matching between these faces and their corresponding expression labels; to do so, we need to define a high discriminative face representation. To find such representation, we followed recent state-of-the-art methods that explore DCNN models to project the face into a new feature space. Through a deep processing, these models extract automatically relevant non-linear features and arrange them into a set of Feature Maps (FMs). Then, we compute a covariance descriptor over these FMs to define a global face representation. In addition, we extract local features by mapping relevant facial regions on the extracted deep FMs to define local covariance descriptors around the eyes, mouth and left/right cheeks.

As a first step, our approach uses a DCNN model to extract deep features that encode well the facial expression in the input face image. In this work, we use the \textit{ExpNet}~\cite{ding2017facenet2expnet} network regularized by the \textit{VGG-face}~\cite{parkhi2015deep} model. 


\subsection{Global DCNN Features}\label{sect:GDCNN}
VGG-face is a DCNN model composed of $16$ layers and trained on $2.6$M facial images for the face identification task. After fine-tuning, VGG-face has also shown competitive performance in recognizing facial expressions. However, given that the model was firstly trained for face recognition on a large dataset, it is expected to still capture facial identity information, especially when it is fine-tuned on a small dataset, like those available for our task. Actually, this identity information should be filtered-out in order to capture person-independent facial expressions. 
Ding \emph{et al.}~\cite{ding2017facenet2expnet} have addressed this issue by proposing the ExpNet model. The architecture of this new model is much smaller than VGG-face, containing only five convolutional layers and one fully connected layer. The key idea is to use VGG-face to regularize this small model in a two-stage training algorithm.

As Ding \emph{et al.} proposed in~\cite{ding2017facenet2expnet}, we first use the target expression dataset to fine-tune the VGG-face model by minimizing the cross-entropy loss. Then, we explore this fine-tuned model to regularize the ExpNet network. Finally, the last convolutional layer of this model is used to extract deep facial features. In what follows, we denote the set of extracted FMs from an input face image $f$ as $\Phi(f)=\{M_1, M_2, \dots, M_{m}\}$, where $\{M_i\}_{i=1}^{m}$ are the $m$ FMs at the last convolutional layer, and $\Phi(.)$ is the non-linear function induced by the DCNN architecture at this layer. 

\subsection{Local DCNN Features}\label{sect:local-DCNN}
In order to explore local information, we extract from the global feature maps $\Phi(f)$ local deep features that are related to relevant facial regions.

To this end, we first detect a set of facial landmarks on the input image. Using these points, four regions $\{R_{j}\}_{j=1}^4$ are identified around the eyes, mouth, and the two cheeks. To localize these facial regions on the FMs, we need to define a pixel-wise mapping between the input face image and its corresponding FMs. 
Actually, a feature map $M_{i}$ results from applying a convolution with linear filters across the input face image. Consequently, units of the feature map will be attached to different facial regions $R_{j}$. Based on this assumption, it is possible to map the coordinates of the feature maps to those of the input face image. Formally, each pixel in the input face image of coordinates $(x_{p}, y_{p})$, can be associated to the feature $\phi_{p}(f, M_{i})$ in the feature map $M_{i}$ such that, 
\begin{equation}
\label{eq:Mapping}
\phi_{p}(f, M_{i}) = M{_i}(\overline{s_1 \times x_{p}}, \overline{s_2\times y_{p}}) \; ,
\end{equation}

\noindent
where $\bar{(.)}$ is the rounding operation, and $s_1$, $s_2$ are the map size ratio with respect to the input size, such that $s_1=\frac{w}{W}$ and $s_2=\frac{h}{H}$, being $w$ and $h$ the width and the height of the feature maps, respectively, and $W$ and $H$ those of the input image. 
Using this pixel-wise mapping, we map each region $R_j$ formed by $r$ pixels $\{p_1,p_2,\dots,p_r\}$ on the input image into the global FMs $\{M_i\}_{i=1}^{m}$ to obtain the corresponding local FMs $\Phi^{R_j}(f) = \{\phi_{p_1}(f, M_{i}),\phi_{p_2}(f, M_{i}),\dots,\phi_{p_r}(f, M_{i})\}_{i=1}^{m}$.

\begin{figure}[!ht]
\centering
\includegraphics[width=\linewidth]{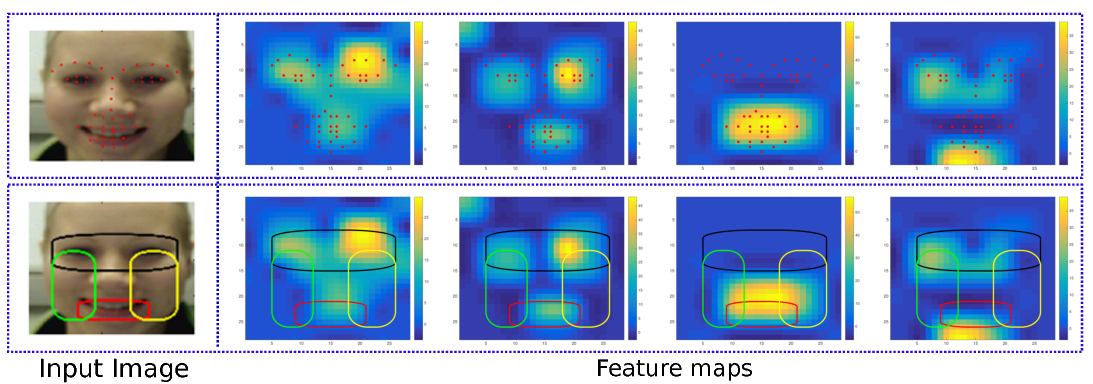}
\caption{Visualization of the detected facial landmarks (first row) and regions (second row) on the corresponding input facial image, and their mapping on four selected feature maps (from $512$) extracted from the last convolution layer of the ExpNet model. Best viewed in color.}
\label{fig:RegionFMs}
\end{figure}

Figure~\ref{fig:RegionFMs} shows the four local regions detected on the input facial image on the left; then, landmarks and regions are shown on four FMs, selected from a total of $512$ FMs.

\subsection{Deep Covariance Descriptors}\label{sect:dcnn-covariance}
Motivated by the impressive performance of covariance matrices as global and local descriptors used in several previous works~\cite{carreira2012semantic,tuzel:2006}, we propose to compute local and global covariance descriptors on the extracted deep features. In particular, a global covariance descriptor is calculated on the global FMs $\Phi(f)$ representing the whole face. In addition, four local covariance descriptors are computed for the four facial regions introduced previously across their corresponding local FMs $\Phi^{R_{j}}(f)$. 
By doing so, we explore a compact and discriminative face representation that encodes all linear correlations between the deep facial features. Contrary to fully connected and softmax layers, this representation allows us to define local descriptors that focus on relevant facial regions. In the following, we describe more formally how to construct the global deep covariance descriptors; the same processing is applied to the local deep covariance descriptors computed over local deep features.

\begin{figure*}[!ht]
\centering\includegraphics[scale=0.15]{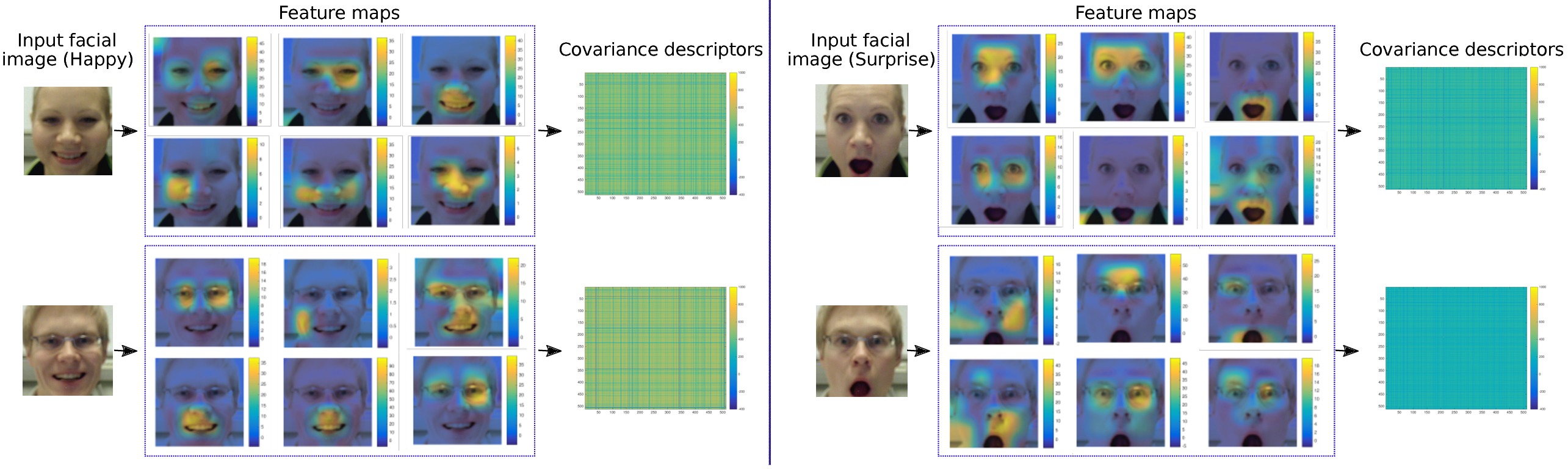}
\caption{Visualization of some feature maps extracted from the last convolution layer of the ExpNet model. The FMs are superimposed on the top of the input image, with their corresponding covariance descriptors for two subjects from the Oulu-CASIA dataset conveying happy and surprise expressions. We show six FMs (selected from $512$ FMs) for each example image. The corresponding covariance descriptors are computed over the $512$ FMs. Best viewed in color.}
\label{fig:feature-maps}
\end{figure*}

The extracted features $\Phi(f)$ are arranged in a $(m \times w \times h)$ tensor, where $w$ and $h$ denote the width and height of the FMs, respectively, and $m$ is their number. Each feature map $M_i$ is vectorized into a $n$-dimensional vector with $n=w \times h$, and the input tensor is transformed to a set of $n$ observations stored in the matrix $[v_{1}, v_{2},\dots,v_{n}] \in \mathbb{R}^{m \times n}$. Each observation $\{v_i\}^n_{i=1} \in \mathbb{R}^{m}$ encodes the values of the pixel $i$ across all the $m$ feature maps. Finally, we compute the corresponding $(m \times m)$ covariance matrix,
\begin{equation}
\label{eq:covariance}
C_{\Phi(f)} = \dfrac{1}{n-1}\sum^{n}_{i=1} (v_{i}-\mu)(v_{i}-\mu)^{T} \; ,
\end{equation}

\noindent
where $\mu=1/n\sum^{n}_{i=1}v_{i}$ is the mean of the feature vectors.

Figure~\ref{fig:feature-maps} shows six selected FMs (chosen from the $512$ FMs extracted with the ExpNet model) for two subjects with happy and surprise expression. The figure also shows the global covariance descriptor relative to the $512$ FMs as a 2D image. Common patterns can be observed in the covariance descriptors computed for similar expressions, \emph{e.g.}, the green color dominates in the covariance descriptors of happy expression (left panel), while the cyan color dominates in the covariance descriptors of surprise expression (right panel).

Covariance matrices of size $m\times m$ are by nature Symmetric Positive Definite (SPD) matrices that are usually studied under a Riemannian structure of the SPD manifold $Sym^{++}(m)$~\cite{tuzel:2006,wang2017discriminative,jayasumana2015kernel}. One of the most used metrics to compare these matrices on $Sym^{++}(m)$, is the Log-Euclidean Riemannian Metric (LERM)~\cite{arsigny2006log}, due to its excellent theoretical properties with simple and fast computation. More formally, the log-Euclidean distance $d_{LERM}: (Sym^{++}(m) \times Sym^{++}(m)) \rightarrow \mathbb{R}^{+}$ between two covariance descriptors $C_{\Phi(f_1)}$ and $C_{\Phi(f_2)}$ of two faces $f_{1}$ and $f_{2}$, is defined by,
\begin{equation}
\label{eq:LERM}
d_{LERM}(C_{\Phi(f_1)},C_{\Phi(f_2)})=\| \log(C_{\Phi(f_1)})-\log(C_{\Phi(f_2)} ) \|_{\textrm{F}} \; ,
\end{equation}

\noindent
where $\|\cdot\|_{\textrm{F}}$ is the Frobenius norm, and $\log(.)$ is the matrix logarithm.

\section{RBF Kernel for Deep Covariance Descriptors Classification of Static Expressions}\label{sect:dcnn-classification}
Considering the geometry of the covariance matrices as points on the non-linear manifold $Sym^{++}(m)$, facial expression classification comes back to the problem of classifying the corresponding covariance descriptors in $Sym^{++}(m)$. To better explore the discriminative ability of these representations, we need to define a suitable classifier that respects their space structure, while standard machine learning techniques cannot be applied directly in such a non-linear space. 
Accordingly, many works proposed adaptations of standard machine learning techniques to the SPD manifold. For example, Harandi \emph{et al.}~\cite{harandi2016sparse} proposed kernels derived from two Bregman matrix divergences, namely, the Stein and Jeffrey divergences to classify SPD matrices in their embedding manifold. 
Here, we benefit from the log-Euclidean distance given by Eq.~\eqref{eq:LERM} between symmetric positive definite matrices to define the Gaussian RBF kernel $K : (Sym^{++}(m) \times Sym^{++}(m)) \rightarrow \mathbb{R}^{+}$,
\begin{equation}
\label{eq:kernel}
K(C_{\Phi(f_1)}, C_{\Phi(f_2)})=\exp (-\gamma d_{LERM}^2(C_{\Phi(f_1)}, C_{\Phi(f_2)})) \; ,
\end{equation}

\noindent
where $d_{LERM}(C_{\Phi(f_1)}, C_{\Phi(f_2)})$ is the log-Euclidean distance between $C_{\Phi(f_1)}$ and $C_{\Phi(f_2)}$. Conveniently for us, this kernel has been already proved to be a positive definite kernel for all $\gamma > 0$~\cite{jayasumana2015kernel}.

\subsection{Fusion of Global and Local Information}\label{sect:fusion}
Each facial region provides relevant information for facial expression analysis and provides a different contribution to the final decision. Consequently, an efficient fusion method of the information provided by different regions is required.

In this section, we investigate different strategies to combine the local information extracted from different facial regions. We divide these strategies into \textit{late fusion} and \textit{early fusion}.
For the late fusion strategy, each region is pre-classified independently, then the final decision is based on the fusion of the scores of the different regions. 
More formally, given $\{\{C^{R_i}_{\Phi(f_j)}\} _{i=1}^4, y_j\}_{j=1}^N$ a set of $N$ training samples for each of the four facial regions with their associated labels, we use Support Vector Machines (SVM) to learn a classifier for each region independently. Each of these classifiers provides for each sample $C_{\Phi(f)}^R$ a scores vector $S^R_{C_{\Phi(f)}} = [s_1, s_2, \dots, s_{l}]$, where $l$ is the number of investigated classes, and $s_i$ is the probability that $C_{\Phi(f)}^R$ belongs to the class $y_i$. Using late fusion, the final scores vector of a sample $C_{\Phi(f)}$ is given by,
\begin{equation}
\label{eq:lateFusionProduct}
S_{C_{\Phi(f)}}=\prod_{i=1}^4 S_{C_{\Phi(f)}}^{R_i} \; ,
\end{equation}

\noindent
for the product rule, and by,
\begin{equation}
\label{eq:lateFusionSum}
S_{C_{\Phi(f)}}=\sum_{i=1}^4 \beta_i S_{C_{\Phi(f)}}^{R_i} \; ,
\end{equation}

\noindent
for the weighted sum rule, where $\beta_i$ represents the weight associated to the region $R_i$.

Concerning the early fusion strategy, we do not need to train a classifier on each region independently; instead, it aims to combine information before any training. A simple way to do so is to concatenate features of all regions in one vector that will be used to train the classifier. This is different from using the global features since many other irrelevant regions are ignored in this case. We refer to this method in our experimental study as \textit{feature fusion}. A more efficient way to conduct early fusion is Multiple Kernel Learning (MKL), where information fusion is performed at the kernel level. In our case, we use MKL to combine different local features using different kernels, such that each kernel $K^R$ is computed on the features of one region $R$ following the weighted sum rule, the final kernel is,
\begin{equation}
\label{eq:EarlySumFusion}
K=\sum_{i=1}^4 \beta_iK^{R_i} \; ,
\end{equation}

\noindent
where $\beta_i$ is the weight associated to the region $R_i$. In what follows, we will refer to the kernel fusion with the weighted sum rule as kernel fusion.

In our experimental study, we have evaluated each of the fusion strategies discussed in this section.

\section{Modeling Dynamic Facial Expressions as Trajectories in $Sym^{++}(m)$}\label{sect:trajectories}
Facial expressions are much more described by a dynamic process than a static one, thus we need to extend our approach to take into account the temporal dimension. To this end, we propose to model a video sequence of a facial expression as a time varying trajectory on the $Sym^{++}(m)$ manifold.

\begin{figure*}
\label{fig:temporal-evolution}
\centering
\includegraphics[width=0.6\linewidth]{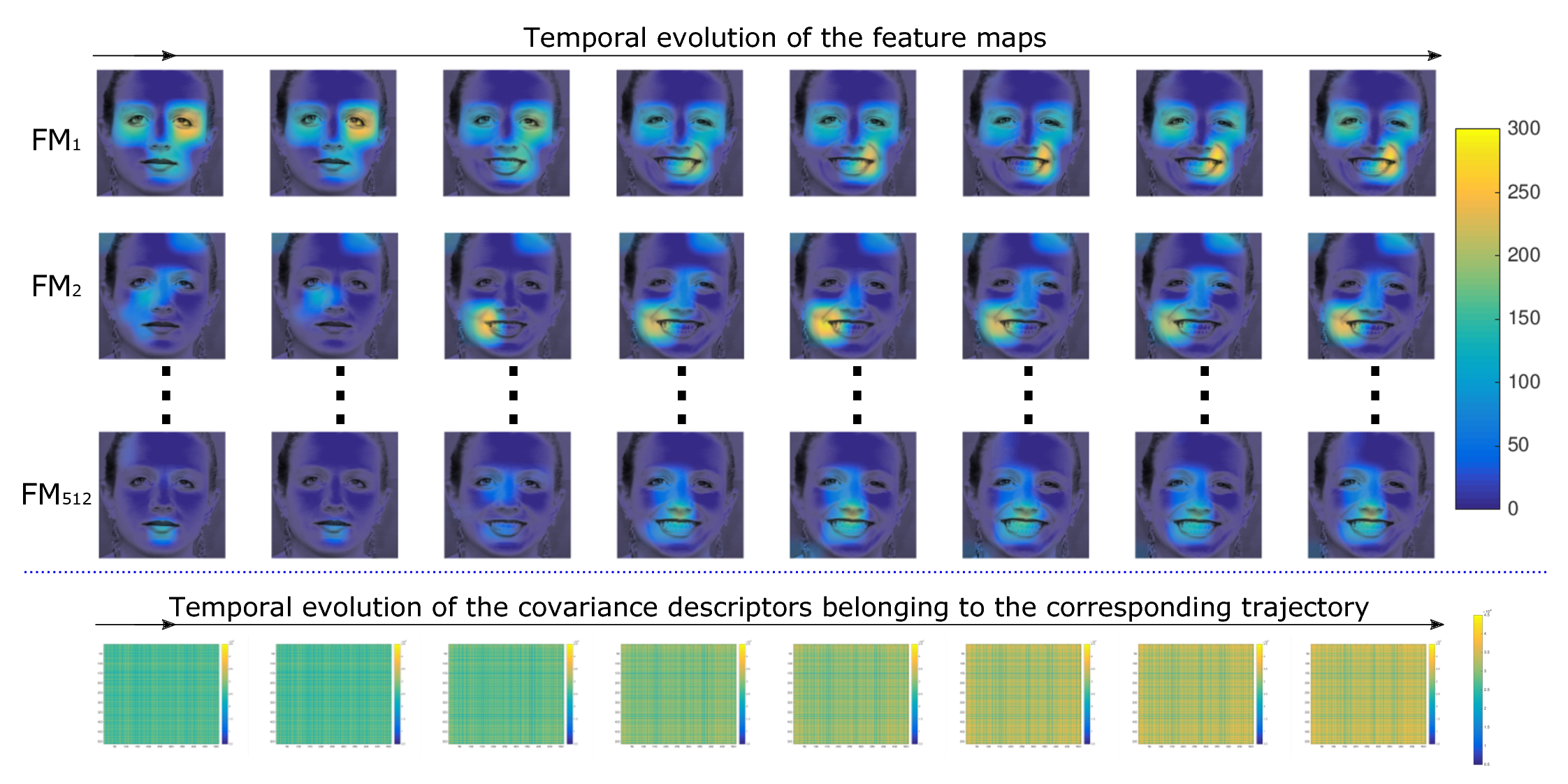}
\caption{Visualization of the temporal evolution of three FMs extracted from the last convolution layer of the ExpNet model. Each row corresponds to the temporal evolution of one FM from $512$. The FMs are superimposed on the top of the input video frame selected from the CK+ dataset. The last row shows the temporal evolution of the corresponding trajectory (sequence of covariance descriptors) of the video. Best viewed in color.}
\label{fig:my_label}
\end{figure*}

Following our static approach, we represent each frame $f$ of a sequence by a covariance matrix $C_{\Phi(f)}$ computed on the top of deep features. Given that each covariance matrix is a point on $Sym^{++}(m)$ as discussed before, a sequence $\{C_{\Phi (f_{i})}\}_{i=1}^{L}$ of $L$ covariance matrices computed on DCNN features defines a trajectory $T_{C_{\Phi}}$ on the $Sym^{++}(m)$ manifold by $T_{C_{\Phi}}: [0,1] \rightarrow Sym^{++}(m)$. We define a trajectory $T_{C_{\phi}}$ to be a path that consists of a set of $L$ points on $Sym^{++}(m)$. 
In Figure~\ref{fig:my_label}, we visualize the temporal evolution of some FMs extracted by our ExpNet model from a normalized video sequence of the CK+ dataset. This figure shows that each FM focuses on some relevant features (related to the facial expression) that are more activated than others over time. For example, the first row (first FM) shows the activation over time of the right mouth corner resulting from the smile movement, while the second FM detects the same activation over time on the left corner. The last row of the same figure illustrates the temporal evolution of the corresponding trajectory. 
In particular, by encoding the $m$ FMs of each frame in a compact covariance matrix, the problem of analyzing the temporal evolution of $m$ FMs is turned to studying a trajectory of covariance matrices in $Sym^{++}(m)$. Here, we can observe that the dominant color of the covariance matrices corresponding to neutral frames is green; this color gradually changes to yellow along the facial expression (\emph{i.e.}, happiness).

Using the same strategy, we extend the local approach as well, by representing each video sequence with five trajectories $\{T_{C_{\Phi}}, \{T_{C_{\Phi}^{Rj}}\}_{j=1}^{4} \}$, including a trajectory which encodes the temporal evolution of the global features, and four trajectories representing the temporal evolution of four facial regions. For simplicity, we will use $T$ to refer to the trajectory $T_{C_{\Phi}}$ in the rest of this section. 

The temporal variability is one of the difficulties encountered when comparing videos. It is due to the different execution rate of the facial expressions, their variable durations, and their arbitrary starting/ending intensities. These aspects yield to a distortion of the comparison measures of the corresponding trajectories. To tackle this problem, different algorithms based on dynamic programming have been introduced to find an optimal alignment between two videos. In this work, we propose to align trajectories in $Sym^{++}(m)$ based on the LERM distance using two algorithms: Dynamic Time Warping (DTW) and Global Alignment (GA).

\subsection{Dynamic Time Warping}
We use the notation of~\cite{cuturi2007kernel} to formulate the problem of aligning trajectories in $Sym^{++}(m)$. Given two trajectories $T^1=\{C^{1}_{\Phi (f_{i})}\}_{i=1}^{L_1}$ and $T^2=\{C^{2}_{\Phi (f_{i})}\}_{i=1}^{L_2}$ of length $L_1$ and $L_2$, respectively, an alignment $\pi$ between these trajectories is a pair of increasing $q$-tuples $(\pi_{1}, \pi_{2})$ of length $q \leq L_1 + L_2 - 1$ such that $1 = \pi_{1}(1) \leq \dots \leq \pi_{1}(q) = L_1$ and $1 = \pi_{2}(1)\leq \dots \leq \pi_{2}(q) = L_2$, with unitary increments and no simultaneous repetitions.

Given $A(T^1,T^2)=\{\pi\}_{i=1}^z$, the set of all $z$ possible alignments between two trajectories $T^1$ and $T^2$, the optimal alignment is given by,
\begin{equation}
\pi^{*} = \operatorname*{argmin}_{\pi \in A(T^1,T^2)} \dfrac{1}{|\pi|} D(\pi) \; ,
\end{equation}

\noindent
where $D(\pi)$, defined as 
\begin{equation}
D(\pi)=\sum_{i=1}^{|\pi|} d( T^1_{\pi_1(i)}, T^2_{\pi_2(i)} ) \; ,
\end{equation}

\noindent
is the cost given by the mean of a local divergence $d$ on $Sym^{++}(m)$ that measures dissimilarities between any two points of the trajectories $T^{1}$ and $T^{2}$. Hence, the dissimilarity measure computed by DTW between $T^1$ and $T^2$ is given by, 
\begin{equation}
D_{dtw}(T^1, T^2)=D(\pi^*) \; .
\end{equation}

To align trajectories in $Sym^{++}(m)$ with DTW, we use the LERM distance $d_{LERM}$ defined in Eq.~\eqref{eq:LERM} to define the divergence $d$. 

The problem of DTW is that the cost function $D_{dtw}$ used for alignment is not a proper metric; it is not even symmetric. Indeed, the optimal alignment of a trajectory $T^1$ to a trajectory $T^2$ is often different from the alignment of $T^2$ to $T^1$. Thus, we can not use it to define a valid positive definite kernel, while the positive definiteness of the kernel is a very important requirement of kernel machines during the classification phase.

\subsection{Global Alignment Kernel}
To address the problem of non positive definiteness of the kernel defined by DTW, Cuturi \emph{et al.}~\cite{cuturi2007kernel} proposed the Global Alignment Kernel (GAK). As shown earlier, DTW uses the minimum value of alignments to align time-series. Instead, the Global Alignment proposes to take advantage of all possible alignments, assuming that the minimum value used in DTW may be sensitive to peculiarities of the time series. GAK has shown its effectiveness in aligning the temporal information in many works including~\cite{lorincz2013emotional, jeni2014spatio,cuturi2011fast}. Furthermore, it requires the same computational effort $\mathcal{O}(L_1L_2)$ as that of DTW. GAK is defined as the sum of exponentiated and sign changed costs of the individual alignments:
\begin{equation}
\label{eq:GAK}
\begin{split}
K_{GA}(T^1, T^2)& = \sum_{\pi \in A(T^1, T^2)} e^{-D(\pi)} \\
 & = \sum_{\pi \in A(T^1, T^2)} \prod_{i}^{|\pi|}e^{-d(T^1_{\pi_1(i)}, T^2_{\pi_2(i)})} \; .
\end{split}
\end{equation}

For simplicity, Eq.~\eqref{eq:GAK} can be rewritten using the local similarity function $k$ induced from the divergence $d$ as $k=e^{-d}$, to get, 
\begin{equation}
\label{eq:GAK2}
   K_{GA}(T^1, T^2) = \sum_{\pi \in A(T^1, T^2)} \prod_{i}^{|\pi|}k(T^1_{\pi_1(i)}, T^2_{\pi_2(i)}) \; .
\end{equation}

\begin{theorem}
\label{theo:KposifivDefinit}
Let $k$ \textit{be a positive definite kernel such that $\frac{k}{k+1}$ is positive definite, then $K_{GA}$ as defined in Eq.~\eqref{eq:GAK2} is positive definite.}
\end{theorem}

According to Theorem~\ref{theo:KposifivDefinit} proved by Cuturi \emph{et al.}~\cite{cuturi2007kernel}, the global alignment kernel $K_{GA}$ is positive definite if $\frac{k}{k+1}$ is positive definite. It has been shown in the same paper~\cite{cuturi2007kernel} that, in practice, most kernels including the RBF kernel satisfy the property that $\frac{k}{k+1}$ provides positive semi-definite matrices. Consequently, in our numerical simulations, we have used the same RBF kernel $K$ given by Eq.~\eqref{eq:kernel} to define our local similarity function $k$. By doing so, we have extended the classification pipeline of our static approach to the dynamic approach by using the same local RBF kernel defined on the $Sym^{++}(m)$ manifold. 
Note that, we checked the positive definiteness of all the kernels used in our experiments. 

\subsection{Classification of Trajectories in $Sym^{++}(m)$}
In this section, we aim to classify the aligned trajectories in $Sym^{++}(m)$. More formally, given a set of aligned trajectories $\mathcal{T}=\{T: [0,1] \rightarrow Sym^{++}(m) \}$, we select a training set $\mathcal{U}=\{ (T^i, Y^i) \}_1^{N_u}$ of $N_u$ samples with their corresponding labels, and we seek for an approximation of the function $g$ that satisfies $Y^i=g(T^i)$ for each sample of the training set $\mathcal{U}$. In order to learn this approximation function, we use two types of SVM, namely, the standard SVM and the pairwise proximity function SVM (ppfSVM)~\cite{gudmundsson2008support}.

Assuming the linear separability of the data, SVM classifies them by defining a separating hyperplane in the data space. However, most of the data do not satisfy this assumption and necessitate to use a kernel function $K$ to transform them to a higher dimensional Hilbert space, where the data are linearly separable. The kernel function can be used with general data types like trajectories. However, according to Mercer's theorem~\cite{shawe2004kernel}, the kernel function must define a symmetric positive semi-definite matrix to be a valid kernel; otherwise, we cannot guarantee the convexity of the resulting optimization problem, which makes it difficult to solve.

Given that GAK provides a valid SPD kernel under a mild condition as demonstrated by Cuturi \emph{et al.}~\cite{cuturi2007kernel}, and given that our local kernel $k$ satisfies this condition as discussed before, we use the standard SVM with the $K_{GA}$ kernel given in Eq.~\eqref{eq:GAK} to classify the aligned trajectories with global alignment on $Sym^{++}(m)$.

By contrast, DTW cannot define a positive definite kernel. Hence, we adopt the algorithm ppfSVM, which assumes that instead of a valid kernel function, all that is available is a proximity function without any restriction. In our case, the proximity function $\mathcal{P} : \mathcal{T} \times \mathcal{T} \rightarrow \mathbb{R}^+$ between two trajectories $T^{1}$ and $T^{2}$ is defined by,
\begin{equation}
\mathcal{P}(T^{1}, T^{2})=D_{dtw}(T^{1}, T^{2}) \; .
\end{equation}

\noindent
Using this proximity function, the main idea of ppfSVM is to represent each training example $T$ with a vector $[\mathcal{P}(T,T^1), \dots , \mathcal{P}(T,T^{N_u})]$, which contains its proximities to all training examples in $\mathcal{U}$. This results in a $N_u \times N_u$ matrix $\Gamma_{\mathcal{U}}$ that contains all proximities between all training data in ${\mathcal{U}}$. Using the linear kernel on this data representation, the kernel matrix $K_{dtw} = \Gamma_{\mathcal{U}} \times \Gamma_{\mathcal{U}}^\mathsf{T}$ is used with SVM to classify trajectories on their manifold. 

Concerning local trajectories, we firstly align them with GAK, then we compute the kernel function given by Eq.~\eqref{eq:GAK2} for each region. Finally, the kernel fusion discussed in Section~\ref{sect:fusion} is used to combine them and classify their corresponding expression.

\section{Experimental results}\label{sect:results}
In this section, we evaluate the effectiveness of our proposed approach in recognizing basic facial expressions. We evaluated the different settings discussed before on several publicly available benchmarks representing constrained and unconstrained environments. 

\subsection{Benchmarks}
We evaluated our approach on the three following datasets:

\textbf{Oulu-CASIA~\cite{zhao2011facial}}: This dataset contains over $480$ videos of $80$ subjects. Each one of these subjects has six videos corresponding to six basic emotion labels; All videos begin with a neutral expression and end with the apex of the corresponding expression. The DCNN model used for this dataset was trained on $1440$ images corresponding to the last three peak frames of each video. These images were also used for the testing of our static approach using a ten-fold cross validation with subject independent splitting. The same setting was conducted for the dynamic approach using all video frames. 

\textbf{Extended Cohn Kanade (CK+)~\cite{lucey2010extended}}: This dataset comprises $327$ sequences of posed expressions, annotated with seven expression labels. Each sequence starts with a neutral expression, and reaches the peak in the last frame. Following the protocol of~\cite{ding2017facenet2expnet}, the three last frames of each sequence are used to represent the video in the static approach, and the subjects are divided into ten groups by ID in ascending order to conduct 10 cross validation.

\textbf{Static Facial Expression in the Wild (SFEW)~\cite{dhall2015video}}: Different from the previous controlled datasets, this database is used  for spontaneous facial expression recognition in the wild. It contains $1,322$ images collected from real movies and labeled with seven facial expressions (Anger, Disgust, Fear, Happiness, Sadness, Surprise and Neutral). It includes three sets: training ($891$ samples), validation ($431$ samples), and test set. Given that we do not have access to the test set labels, all the results of this dataset in our experiments were reported on the validation set.

\textbf{Acted Facial
Expressions in the Wild (AFEW 6.0)~\cite{dhall2012collecting}}:
It is a dynamic non-controlled dataset that contains videos selected from movies. It is composed of 1,156 labeled videos of which 773 samples are used for training and 383 for validation. This dataset contains the same seven facial expressions categories as SFEW. The results are reported on the validation set of this dataset since we do not have access to the test set.

\subsection{Settings}
As data processing, we first applied the Viola \& Jones face detector~\cite{viola2004robust} to the CK+ and Oulu-CASIA datasets. Concerning SFEW, we utilized the aligned faces provided with the database. Then, we used the Chehra face tracker~\cite{asthana2014incremental} to localize $49$ facial landmarks explored in the local approach to extract facial regions. Concerning AFEW dataset, we used OpenFace\footnote[1]{https://github.com/TadasBaltrusaitis/OpenFace/wiki} for landmarks detection and face alignment. All the detected faces were cropped and resized to $224 \times 224$ to be fed to the DCNN model. For the dynamic approach, we firstly normalize videos using the method proposed by Zhou \emph{et al.}~\cite{zhou2011towards}.

\textbf{DCNN models training}: In order to keep our experiments consistent with the state-of-the-art~\cite{ding2017facenet2expnet, ofodile2017automatic}, we trained a DCNN model for each dataset separately. For CK+ and Oulu-CASIA, the training was done in ten cross validation, which results in ten DCNN models (one model per fold) for each dataset; each one of these models was trained on nine splits and tested on the rest split. Since SFEW is divided on training and validation sets, we trained its corresponding DCNN model on the training set. The model used for AFEW was trained on SFEW dataset that contains its static peak frames. Following~\cite{ding2017facenet2expnet} and~\cite{ofodile2017automatic}, we performed the training of all these models in two steps:
\begin{itemize}
\item \textbf{VGG-face fine-tuning} -- As first step, we fine-tuned the VGG-face model~\cite{parkhi2015deep} on our datasets. The training was performed in 100 epochs adopting Stochastic Gradient Descent as optimization algorithm. The mini-batch size was fixed to $64$, the momentum to $0.9$, and the learning rate to $0.0001$ decreased by $0.1$ after each $50$ epochs. The horizontal flipping of the original data was used for data augmentation, and a Gaussian distribution was utilized to initialize the fully connected layers that were trained from scratch with the appropriate number of classes. 
\item \textbf{ExpNet training} -- The ExpNet architecture is composed of five convolutional layers, each one followed by ReLU activation and max pooling layer. The ExpNet training was done in two steps; we firstly train the convolutional layers that were regularized with our fine-tuned VGG-face models for 50 epochs; then, we append one fully connected layer of 128 neurons to train the whole network for additional 50 epochs. For more details about the ExpNet architecture and all the training parameters (learning rate, momentum, mini-batch size, etc), the reader is referred to~\cite{ding2017facenet2expnet}. All the training experiments were conducted with the deep learning framework Caffe~\cite{jia2014caffe}.
\end{itemize}

\textbf{Feature extraction}: Given that the last pooling layer is the nearest one to the classification layers (fully connected and softmax layers), it is natural that it provides the most discriminating features. Based on this motivation, we chose to extract the deep features of each face from the last pooling layer. The features extracted from this layer are organized as $512$ FMs of size $7 \times 7$, which results in covariance descriptors of size $512 \times 512$ according to Eq.~\eqref{eq:covariance}. 
For the local treatment, we first used the detected landmarks to localize the facial regions (eyes, mouth and the two cheeks) on the input image. Then, we mapped these regions to the FMs using Eq.~\eqref{eq:Mapping} with a ratio of $s_1=s_2=1/16$. Note that, before the mapping, we re-sized all the FMs to $14 \times 14$, which allows us to better map landmarks from the input image coordinates to the FM coordinates and minimize the overlapping between the facial regions. The local features extracted around each region are explored to compute local deep covariance descriptors of size $512 \times 512$. According to Eq.~\eqref{eq:covariance}, despite the different sizes of the extracted regions, the resulting covariance descriptors have the same size (depending only on the FMs number) lying in the same SPD manifold $Sym^{++}(512)$. Figure~\ref{fig:feature-maps} shows some FMs extracted with the last pooling layer of the ExpNet model and their corresponding covariance descriptors.

\textbf{Image Classification}: Each static face image was represented by a covariance descriptor of size $512 \times 512$ in the global approach, and by four local covariance descriptors in the local approach. To efficiently compare these descriptors in their manifold $Sym^{++}(512)$, it is empirically necessary to ensure their positive definiteness by using their regularized version, $C_{\Phi(f)} + \epsilon I$, where $\epsilon$ is a regularization parameter (set to $0.0001$ in all our experiments), and $I$ is the $512 \times 512$ identity matrix. The classification of these static descriptors was conducted using multi-class SVM with Gaussian kernel on $Sym^{++}(512)$. The parameters involved by SVM and the Gaussian kernel as well as those used for the fusion methods that require weights, were set using cross validation with grid search. To note that, except Table~\ref{tab:Accuracy}, all the results reported here are obtained using the ExpNet model since it provides better results \emph{w.r.t.} the VGG-face model according to Table~\ref{tab:Accuracy}. 

For the dynamic datasets (Oulu-CASIA and CK+), we followed the setting of Ding \emph{et al.}~\cite{ding2017facenet2expnet}. Accordingly, each video was represented by its last three peak frames and the distance between two videos was computed as the mean of the distances between their three last frames. In Table~\ref{tab:Accuracy}, we considered a video as correctly classified by the softmax layer if its three last frames were correctly recognized.

\textbf{Video Classification}: For the dynamic approach on CK+ and Oulu-CASIA, each video was represented as a trajectory of $15$ points in $Sym^{++}(512)$ and by four local trajectories of $15$ points for the local approach, where each point is a regularized covariance matrix of size $512 \times 512$. Given that the videos of the AFEW dataset contain more frames than the other datasets, we chose to normalize its videos to 30 frames. Accordingly, the trajectories of this dataset are composed of 30 points in $Sym^{++}(512)$. These trajectories were aligned and classified with SVM using the kernel functions discussed earlier. The fusion of local trajectories was performed with kernel fusion, which has shown the best results in the static approach. 

\subsection{Results and Discussion}
\subsubsection{Static Facial Expressions}
As first analysis, we investigate the performance of using covariance descriptors to encode global (G-FMs) and local (R-FMs) deep features. To this end, we compare in Table~\ref{tab:Accuracy} the results of our approach with those obtained with classical DCNN classification (\emph{i.e.}, fully connected and softmax layers) using two DCNN models, VGG-face and ExpNet. We did not include AFEW dataset in this table since, in contrast to CK+ and Oulu-CASIA, we can not localize the peak frames in its videos.

On Oulu-CASIA, the table shows that the G-FMs solution improves the results of standard classification of the VGG-face and ExpNet models with $3.7\%$ and $1.26\%$, respectively. More improvement is observed on CK+, where it reaches $7.16\%$ and $6.69\%$ for the VGG-face and ExpNet models, respectively.
Though less marked, a gain of $0.92\%$ for ExpNet and $0.69\%$ for VGG-face has been also achieved on SFEW. According to these results, we conclude that encoding linear correlation of the deep features in covariance descriptors yields more effective and discriminative representations. Moreover, our results show that, even if the fully connected and softmax layers were trained in an end-to-end manner with the other layers of the model, the classification of deep covariance descriptors using a Gaussian kernel on the SPD manifold is more effective. 
Table~\ref{tab:Accuracy} also shows that combining local (R-FMs) and global features (G-FMs) attains a clear superiority on the Oulu-CASIA and CK+ datasets outperforming the global method (G-FMs) by $1.25\%$ and $1.33\%$, respectively. By contrast, local features do not show any improvement on SFEW. This can be explained by the failure of facial landmark detection in many cases on this challenging dataset (some failure cases of landmark detection on this dataset are shown in Figure~\ref{fig:landfail}), while our local method requires an accurate detection of the facial landmarks to correctly extract local deep features.

\setlength{\tabcolsep}{4pt}
\begin{table}
\caption{Comparison of the proposed classification scheme (global (G-FMs), and global plus local (G-FMs and R-FMs)) with respect to the VGG-Face and ExpNet models with fully connected layer and softmax\label{tab:Accuracy}}
\centering
\small
\scalebox{0.9}{
\begin{tabular}{llccc}
\hline
\textbf{Dataset} & \textbf{Model} & \textbf{FC-Softmax} & \textbf{G-FMs} & \textbf{G-FMs and R-FMs} \\
\hline
\it Oulu-CASIA & \it VGG Face & 77.8 & 81.5 & -- \\
& \it ExpNet & 82.29 & \textbf{83.55} & \textbf{87.08} \\
\hline
\it CK+ & \it VGG Face & 83.74 & 90.90 & -- \\
& \it ExpNet & 90.38 & \textbf{97.07} & \textbf{98.40} \\
\hline
\it SFEW &\it VGG Face & 46.66 & 47.35 & -- \\
& \it ExpNet & 48.26 & \textbf{49.18} & \textbf{49.18} \\
\hline
\end{tabular}}
\end{table}
\setlength{\tabcolsep}{1.4pt}

Table~\ref{tab:FusionsComparison} compares the fusion modalities discussed in Section~\ref{sect:fusion}. We found consistent results across the datasets, indicating the kernel fusion and weighted sum late fusion are the best methods to combine local and global covariance descriptors. 

We investigated in Table~\ref{tab:ResultsRegions}, the contribution of each facial region used in our method in recognizing the corresponding facial expression. According to this table, the eye region is the best performing facial region on CK+ and Oulu-CASIA. By contrast, on SFEW and AFEW the eye region does not achieve good performance. As previously discussed, this can be motivated by the less accurate landmark detection in non-frontal views and the occlusions that are usually encountered in in-the-wild environment, which badly affects the localization of the region and its corresponding deep features. Concerning the rest regions, the right and left cheeks show almost the same score surpassing with a large gain the mouth score. On all the datasets, the mouth region provides generally the worst score. We may explain this result by the small size of this region \emph{w.r.t.} the other regions. Hence, the mouth region is usually represented by a small number of deep features (sometimes 4 or 8 features), while the other regions are represented by a larger number of features.

\setlength{\tabcolsep}{4pt}
\begin{table}[!ht]
\caption{Overall accuracy (\%) of different fusion schemes on the Oulu-CASIA, CK+, and SFEW datasets. Results of early fusion methods are reported in the first group followed by results of late fusion methods in the second group}
\label{tab:FusionsComparison}
\begin{center}
\small
\begin{tabular}{lccc}
\hline
\textbf{Fusion method} & \textbf{Oulu-CASIA} & \textbf{CK+} &\textbf{SFEW} \\
\hline
\it Features fusion (R-FMs only) & 84.38  &  96.70 & 45.70  \\
\it Kernels fusion  & \textbf{87.08}  & 98.28 & 48.72 \\
\hline
\it Weighted-sum fusion  &  84.80 & \textbf{98.40} & \textbf{49.18}  \\
\it Product fusion & 84.05 & 96.41 & 45.24  \\
\hline
\end{tabular}
\end{center}
\end{table}

\setlength{\tabcolsep}{4pt}
\begin{table}[!ht]
\caption{Overall accuracy (\%) of different regions and the best fusion results on the Oulu-CASIA, SFEW, and CK+ datasets for the ExpNet model}
\label{tab:ResultsRegions}
\centering
\small
\begin{tabular}{lcccc}
\hline
\textbf{Region}  & \textbf{Oulu-CASIA}  & \textbf{CK+} & \textbf{SFEW} & \textbf{AFEW}  \\
\hline
\it Eyes & 84.59  & 93.47 & 38.05& 40.32 \\
\it Mouth & 70.00 & 83.34 & 38.98 & 37.60\\
\it Right Cheek&  83.96 & 84.56 & 43.16 &  42.23\\
\it Left Cheek &  83.12 & 83.61 & 42.93 & 43.32 \\
\hline
\it R-FMs fusion & 86.25 & 98.28 & 45.70 & 46.04  \\
\it G-FMs and R-FMs fusion & 87.08 &  98.40 &  49.18 & 49.59 \\
\hline
\end{tabular}
\end{table}
\setlength{\tabcolsep}{1.4pt}

\subsubsection{Dynamic Facial Expressions}
In Table~\ref{tab:dynamic}, we report results of the dynamic approach on CK+ and Oulu-CASIA, using either GAK with SVM or DTW with ppfSVM to align and classify the deep trajectories. We divide the methods into two groups: the first group uses global covariance descriptors (G-Traj); the second group corresponds to the fusion of local covariance trajectories (R-Traj). Unsurprisingly, on all the datasets, GAK achieved the highest accuracy compared with DTW. On CK+, GAK achieved an improvement of $4.62\%$ and $3.12\%$, with global trajectories \textit{G-FMS} and local trajectories \textit{R-FMS}, respectively. On the other hand, this improvement reaches about $4.12\%$ and $2.94\%$, with \textit{G-FMS} and \textit{R-FMS}, respectively, on Oulu-CASIA. In consistency with this results, GAK improved the results on the AFEW in-the-wild-dataset by $5.16\%$ and $6.54\%$ \emph{w.r.t.} DTW for global and local trajectories, respectively.
These results indicate the effectiveness of the proposed global alignment with RBF kernel on $Sym^{++}(m)$ in classifying trajectories on their SPD manifold; they also show the importance of using a symmetric positive definite kernel instead of the pairwise proximity function used with DTW. The same table shows consistent results with those of the static approach, where the fusion of the local trajectories surpasses the performance of the global trajectory by $3.83\%$ on CK+, $3.79\%$ on Oulu-CASIA and $3.27\%$ on AFEW, using GAK. This improvement is also observed with DTW by $5.33\%$ on CK+, $4.97\%$ on Oulu-CASIA and $1.91\%$ on AFEW, which confirms the contribution of the local analysis of facial expressions.
We notice that the degradation observed between the static and dynamic approaches on CK+ and Oulu-CASIA datasets can be explained by many factors, among them the fact that video classification is more challenging taking into account the temporal evolution and its challenges. Furthermore, for the dynamic approach, the video contains intermediate frames, which do not correspond to any facial expression. Such frames have not been used during the training of DCNN models. Thus, it is not surprising that the DCNN model can perform worse on the intermediate frames of the video.


\setlength{\tabcolsep}{4pt}
\begin{table}[!ht]
\caption{Overall accuracy (\%) of different dynamic methods on CK+ and Oulu-CASIA. Results based on global covariance trajectories (\textit{G-Traj}) are reported in the first group, followed by the results of the fusion of region covariance trajectories (\textit{R-Traj}) in the second group. Kernel fusion is adopted here as fusion method}
\label{tab:dynamic}
\centering
\small
\begin{tabular}{lccc}
\hline
\textbf{Method} & \textbf{Oulu-CASIA} & \textbf{CK+} & \textbf{AFEW} \\
\hline
\it G-Traj + DTW + ppfSVM & 78.13 & 89.71 &  41.14\\
\it \textbf{G-Traj + GAK + SVM} & \textbf{82.25} &  \textbf{94.33} &  \textbf{46.32} \\
\hline
\it R-Traj + DTW + ppfSVM &  83.10 & 95.04 & 43.05\\
\it \textbf{R-Traj + GAK + SVM}  &  \textbf{86.04} &  \textbf{98.16} & \textbf{49.59}\\
\hline
\end{tabular}
\end{table}
\setlength{\tabcolsep}{1.4pt}

\setlength{\tabcolsep}{4pt}
\begin{table}[!ht]
\caption{Comparison with state-of-the-art solutions on CK+. Geometric, appearance, and hybrid solutions are reported in the first three groups of methods, respectively; Our solutions are given in the last two rows}
\label{tab:ResultsComparisonsCK}
\begin{center}
\small
\begin{tabular}{lccl}
\hline
\textbf{Method} & \textbf{Accuracy} & \textbf{\# classes} & \textbf{D/S} \\
\hline
\it Taheri et al.~\cite{Taheri:2011} & 85.8 & 7 & Dynamic \\
\it Jung et al.~\cite{jung2015joint} & 92.35 & 7 & Dynamic\\
\it Kacem et al.~\cite{kacem:2017} & 96.87 & 7 & Dynamic \\
\hline 
\it Liu et al.~\cite{liu2013aware} & 92.22 & 8 &  Static\\
\it Liu et al.~\cite{liu2014deeply} & 92.4 & 7 & Dynamic \\
\it  Liu et al.~\cite{liu2014learning} & 94.19 & 7 & Dynamic \\
\it Cai et al.~\cite{cai2018island} & 94.35& 7 & Static \\
\it Meng et al.~\cite{meng2017identity} & 95.37 & 7 & static\\
\it Li et al.~\cite{li2017reliable} & 95.78& 6 & static \\
\it Chu et al.~\cite{chu2017selective}& 96.40 & 7 & Dynamic\\

\it Yang et al.~\cite{yang2018identity} & 96.57 & 7 & Static \\

\it Ding et al.~\cite{ding2017facenet2expnet} & 96.8 & 8 & Static \\
\it Mollahosseini et al.~\cite{mollahosseini2016going} & 97.80 & 7 & Static\\
\it Zhao et al.~\cite{zhao2016peak}  & 97.30 & 6 & Dynamic \\

\it Yang et al.~\cite{yang2018facial} & 97.30 & 7 & Static\\

\it Ding et al.~\cite{ding2017facenet2expnet} & 98.60 & 6 & Static \\
\hline 
\it Jung et al.~\cite{jung2015joint} & 97.25 & 7 & Dynamic \\
\it Ofodile et al.~\cite{ofodile2017automatic}& 98.70 & 7 & Dynamic\\
\hline
\it \textbf{ours (ExpNet + G-FMs)} & \textbf{97.07} & 7 & Static\\
\it \textbf{ours (ExpNet + fusion)} & \textbf{98.40} &
7 & Static \\
\hline
\it \textbf{ours (ExpNet + G-FMs)} & \textbf{94.33} & 7 & Dynamic\\
\it \textbf{ours (ExpNet + fusion)} & \textbf{\textbf{98.16}} &
7 & Dynamic \\
\hline
\end{tabular}
\end{center}
\end{table}

\setlength{\tabcolsep}{4pt}
\begin{table}[!ht]
\caption{Comparison with state-of-the-art solutions on Oulu-CASIA. Geometric, appearance and hybrid solutions are reported in the first three groups of methods; Our solutions are given in the last row}
\label{tab:ResultsComparisonsCASIA}
\begin{center}
\small
\begin{tabular}{lccl}
\hline
\textbf{Method} & \textbf{Accuracy} & \textbf{\# classes} &\textbf{D/S} \\
\hline
\it Jung et al.~\cite{jung2015joint} & 74.17 & 6 & Dynamic\\
\it Kacem et al.~\cite{kacem:2017} & 83.13 & 6 & Dynamic\\
\hline 
\it  Liu et al.~\cite{liu2014learning} & 74.59 & 6 & Dynamic \\
\it Guo et al.~\cite{guo2012dynamic} & 75.52 & 6 & Dynamic\\
\it Cai et al.~\cite{cai2018island} & 77.29 & 6 & Static\\
\it Ding et al.~\cite{ding2017facenet2expnet} & 82.29 & 6 & Static \\
\it Zhao et al.~\cite{zhao2016peak} & 84.59 & 6 & Dynamic \\
\hline 
\it Jung et al.~\cite{jung2015joint} & 81.46 & 6 & Dynamic \\
\it Yang et al.~\cite{yang2018facial} & 88.0 & 6 & Static \\
\it Yang et al.~\cite{yang2018identity} & 88.92 & 6 & Static \\
\it Ofodile et al.~\cite{ofodile2017automatic} & 89.60 & 6 & Dynamic \\
\hline
\it \textbf{ours (ExpNet + G-FMs)} & \textbf{83.55} & 6 & Static \\
\it \textbf{ours (ExpNet + fusion)} & \textbf{87.08} & 6 & Static \\
\hline
\it \textbf{ours (ExpNet + G-FMs)} & \textbf{82.25} & 6 & Dynamic \\
\it \textbf{ours (ExpNet + fusion)} & \textbf{86.04} & 6 & Dynamic \\
\hline
\end{tabular}
\end{center}
\end{table}

\subsubsection{Comparison with the State-of-the-Art}
The performance of several state-of-the-art approaches and that of our static and dynamic methods on CK+, Oulu-CASIA, SFEW and AFEW are given in Table~\ref{tab:ResultsComparisonsCK},~\ref{tab:ResultsComparisonsCASIA}, and~\ref{tab:ResultsComparisonsSFEW},~\ref{tab:ResultsComparisonsAFEW}, respectively. In general, both our static and dynamic solutions achieved competitive performance \emph{w.r.t.} the most recent approaches. 
Comparing the static approaches on CK+ and Oulu-CASIA (Table~\ref{tab:ResultsComparisonsCK} and~\ref{tab:ResultsComparisonsCASIA}, respectively), our method outperforms the state-of-the-art with a significant gain. The method by Ding \emph{et al.}~\cite{ding2017facenet2expnet} outperforms our results on CK+ with an accuracy of $98.60\%$; however, this result is reported on $6$ facial expressions only, ignoring the challenging contempt expression of this database. The approaches proposed in~\cite{yang2018facial} and~\cite{yang2018identity} outperform our static method on Oulu-CASIA, while our results surpass them on CK+.  Concerning the dynamic approaches, we obtained the second highest accuracy on both CK+ and Oulu-CASIA datasets, outperforming several recent approaches. The best accuracy on both datasets are reported by Ofodile \emph{et al.}~\cite{ofodile2017automatic}; however, the details of the frames used in the training of their DCNN model, that are needed to effectively compare the two approaches are not reported in their work. It is worth noting that in order to better compare our static results with those of Ding \emph{et al.}~\cite{ding2017facenet2expnet} on the Oulu-CASIA dataset, we reproduce the performance of their method also on a per-video basis, classifying a video as accurately recognized when its three last peak frames are correctly classified. 

Although the multiple challenges imposed by the SFEW in-the-wild dataset, our static method outperforms various state-of-the-art approaches with a significant gain. 
In Table~\ref{tab:ResultsComparisonsSFEW}, we did not include the approaches that use additional datasets to train their DCNN model. For example, Yu \emph{et al.}~\cite{yu2015image} (55.96\%) use the FER2013 dataset~\cite{goodfellow2013challenges} that provides more than $35,000$ samples to train their DCNN model. In their work, Ding \emph{et al.}~\cite{ding2017facenet2expnet} show that this data augmentation can boost results on SFEW by $6.86\%$.
The same strategy was used in~\cite{acharya2018covariance}, where the model was pre-trained on a subset of an additional dataset (MS-Celeb-1M), while our model was trained only on the training set of the SFEW dataset. We also did not include some works that were conducted in different setting conditions than ours. For example, Kaya \emph{et al.}~\cite{kaya2015contrasting} have reported their results ($53.06\%$) only on $343$ out of $436$ images in the SFEW dataset due to their data alignment algorithms as explained in Section 4 of their paper, while their performance on the 427 images is only $42.15\%$. Kim \emph{et al.}~\cite{kim2015hierarchical} have obtained $53.9\%$ using 216 DCNN models, while we only use a single model.

Regarding the AFEW dataset, Table~\ref{tab:ResultsComparisonsAFEW} shows that our results on this challenging dataset are competitive with the state-of-the-art. In this table, we reach the third highest accuracy after the two approaches that combine multiple DCNN models, while we use just a single model. We note that our results were not compared with the methods that also employ audio features (\emph{e.g.},~\cite{ding2016audio}, 51.20\%; \cite{fan2016video}, 51.96\%; \cite{yao2016holonet}, 51.96\%; \cite{kaya2015contrasting}, 58.22\%). 
It is worth nothing that the results of~\cite{liu2014combining} were reported on AFEW 4.0. Their results reported in Table~\ref{tab:ResultsComparisonsAFEW} on AFEW 6.0 are given according to~\cite{acharya2018covariance}.

\subsection{Challenge encountered with in-the-wild datasets.}
\begin{figure}[!ht]
\centering
\includegraphics[width=\linewidth]{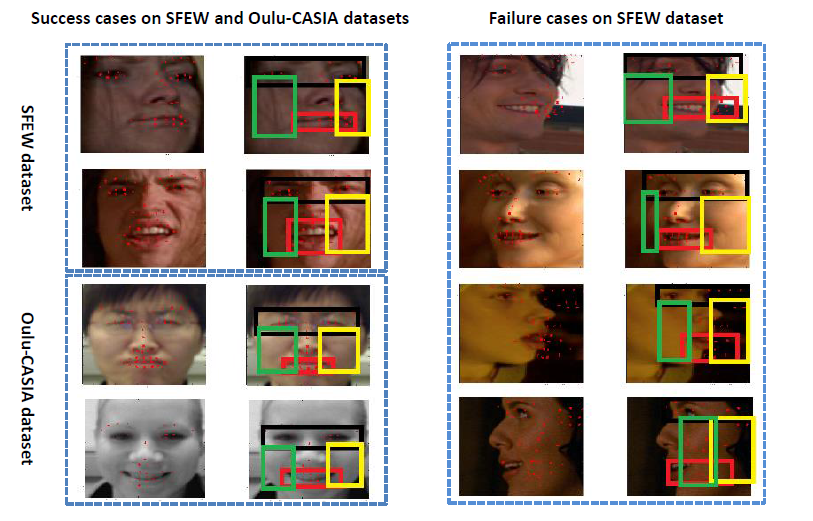}
\caption{Examples of facial landmark and region detection on the SFEW and Oulu-CASIA datasets, with some failure cases for the SFEW dataset. For each example, the image on the left shows the aligned face with its landmark points, while the image on the right represents the aligned face with its detected regions.}
\label{fig:landfail}
\end{figure}

When applied to in-the-wild datasets, our local approach is greatly affected by the performance of the landmark detector. Due to occlusions, non-frontal views and small size of the face in the images of these datasets, it is often more challenging to accurately localize different landmarks, while our local approach relies on the landmarks position to extract the features related to each region. For example, Figure~\ref{fig:landfail} shows some failure and success cases of facial landmark and region detection on the input facial images. In the left panel of this figure, we show examples from the Oulu-CASIA and SFEW datasets, where the landmark and region detection succeeded. In the right panel, we show four failure examples for landmark and region detection in the SFEW dataset. We noticed that this step failed on $\sim30\%$ of the facial images of SFEW. This explains why we do not obtain improvements by combining local and global covariance descriptors on this dataset.

Despite this limitation and according to Table~\ref{tab:ResultsComparisonsSFEW} and~\ref{tab:ResultsComparisonsAFEW}, our method is very competitive with respect to the state-of-the-art and outperforms many recent works even when applied to in-the-wild datasets (\emph{i.e.}, SFEW and AFEW). On the one hand, the results on the SFEW dataset after the fusion of local features did not harm the overall performance since the global features are maintained in all the fusion schemes. On the other, we obtained an improvement of more than $3\%$ on the AFEW dataset when employing the fusion of local and global features.

\setlength{\tabcolsep}{4pt}
\begin{table}[!ht]
\caption{Comparison with state-of-the-art solutions on the SFEW dataset. Our solutions are given in the last row, for (ExpNet + fusion) we have reported the results of the best fusion method using $G-FMs$ and $R-FMs$. The results of the approaches marked with $^+$ are using additional datasets during the training of their model.}
\label{tab:ResultsComparisonsSFEW}
\begin{center}
\small
\begin{tabular}{lc}
\hline
\textbf{Method} & \textbf{Accuracy} \\
\hline
\it Liu et al.~\cite{liu2013aware} & 26.14 \\
\it Levi et al.~\cite{levi2015emotion}& 41.92 \\
\it Kaya et al.~\cite{kaya2015contrasting} & 42.84 \\
\it Mollahosseini et al.~\cite{mollahosseini2016going} & 47.70 \\
\it Ding et al.~\cite{ding2017facenet2expnet}  & 48.29  \\
\it Ng et al.~\cite{ng2015deep} & 48.50 \\
\it Cai et al.~\cite{cai2018island}& 52.52 \\
\it Bargal et al.~\cite{bargal2016emotion} & 59.42 \\
\it Acharya et al.~\cite{acharya2018covariance}$^+$ & 58.14 \\

\hline
\it \textbf{ours (ExpNet + G-FMs)} & \textbf{49.18} \\
\it \textbf{ours (ExpNet + fusion)} & \textbf{49.18} \\
\hline
\end{tabular}
\end{center}
\end{table}


\setlength{\tabcolsep}{4pt}
\begin{table}[!ht]
\caption{Comparison with state-of-the-art solutions on the validation set of the AFEW 6.0 dataset following EmotiW 2016. The results of the methods marked with $^*$ were obtained by fusion of multiple deep models. Our solutions are given in the last row, for (ExpNet + fusion) we have reported the results of the best fusion method using $G-FMs$ and $R-FMs$}
\label{tab:ResultsComparisonsAFEW}
\begin{center}
\small
\begin{tabular}{lc}
\hline
\textbf{Method} & \textbf{Accuracy} \\
\hline
\it Baseline (provided by EmotiW organizers)~\cite{dhall:2016} & 40.47 \\
\it Yan et al.~\cite{yan2018multi} & 44.46 \\
\it Single Best CNN-RNN~\cite{fan2016video} & 45.30\\
\it  Single Best C3D~\cite{fan2016video} & 39.69 \\
\it Single Best HoloNet~\cite{yan2018multi} & 44.57 \\
\it Baseline (RBF Kernel)~\cite{liu2014combining} & 45.95 \\
\it Baseline (Poly Kernel)~\cite{liu2014combining} & 45.43 \\
\it Acharya et al.~\cite{acharya2018covariance} & 46.71 \\
\hline
\it Multiple CNN-RNN and C3D~\cite{fan2016video}$^*$ &  51.80 \\
\it VGG13+VGG16+ResNet~\cite{bargal2016emotion}$^*$   & 59.16 \\
\hline
\it \textbf{ours (ExpNet + G-FMs)} & \textbf{46.32} \\
\it \textbf{ours (ExpNet + fusion)} & \textbf{49.59} \\
\hline
\end{tabular}
\end{center}
\end{table}

\section{Conclusions}\label{sect:conclusion}
In this paper, we proposed deep covariance descriptors and deep covariance trajectories for facial expression recognition from static and dynamic data, respectively. The idea consists of encoding global and local DCNN features in compact covariance matrices. 

A DCNN model trained for facial expression recognition is able to automatically characterize the relevant patterns specific to each facial expression; these patterns are usually related to Facial Action Units~\cite{khorrami2015deep}. In the general approach, the classification of these features is performed by using fully connected layers to flatten these features, then a softmax layer is explored to get a probability for each facial expression. By contrast, in this work, we encode all linear correlations between deep facial features extracted from the last convolutional layer in compact covariance matrices. To respect the nonlinear structure of covariance matrices as points on the SPD manifold, we classified these static descriptors using SVM with a Gaussian kernel defined on SPD manifold. 
Our results show that this classification method is more effective than the standard classification with fully connected and softmax layers. Furthermore, we have shown how our approach can deal with the temporal dynamics of the face. This is achieved by modeling a facial expression video sequence as a deep trajectory in the SPD manifold. To jointly align and classify deep trajectories in the SPD manifold, while respecting the structure of the manifold, a global alignment kernel is derived from the Gaussian kernel, which was used to classify static covariance descriptors. This yields a valid positive definite kernel that is fed to SVM for the final classification of the trajectories. 
By conducting extensive experiments on the Oulu-CASIA, CK+, SFEW and AFEW datasets, we have shown that the proposed approach achieves state-of-the-art performance for facial expression recognition. 

As future work, we aim to train our method in an end-to-end manner to further boost the performance. In this direction, Acharya et al.~\cite{acharya2018covariance} have proposed a network trained in an end-to-end manner that computes covariance descriptors on the convolutional features. This paper exploits the SPD manifold network proposed in~\cite{huang2017riemannian} to conduct an end-to-end training. By contrast, our approach relies on SVM with a Gaussian kernel computed in the SPD manifold and takes advantage of local features, which differ from~\cite{acharya2018covariance}. 
Inspiring solutions for designing an end-to-end network in our case are given in~\cite{wang2017discriminative,huang2017riemannian,Ionescu:2015}.

\section*{Acknowledgements}
This work was supported by the scholarship of Excellence from the National Center for Scientific and Technical Research (CNRST) of Morocco, and by CAMPUS FRANCE [PHC TOUBKAL 2019 (French-Morocco Bilateral Program)] under Grant 41539RH.  It was partially supported by the French State, managed by the National Agency for Research (ANR) under the Investments for the future program with reference ANR-16-IDEX-0004 ULNE.




\ifCLASSOPTIONcaptionsoff
  \newpage
\fi





\vspace{-1.2 cm}
\begin{IEEEbiography}[{\includegraphics[width=1in,height=1.0in,clip,,keepaspectratio]{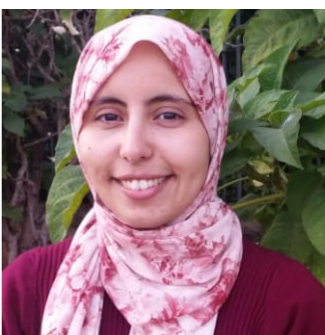}}]{Naima Otberdout} received her master's degree in computer sciences and telecommunication in 2016, from Mohammed V University in Morocco.
Currently, she is a Ph.D. candidate in Computer Science at the same University. Her general research interest is computer vision and machine learning. Particularly, she is interested in deep learning with Riemannian geometry for face analysis and human behavior understanding.
\end{IEEEbiography}

\vskip -2\baselineskip plus -1fil

\begin{IEEEbiography}[{\includegraphics[width=1in,height=1.20in,clip,keepaspectratio]{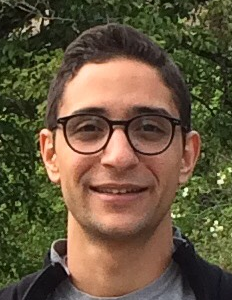}}]{Anis Kacem} is a Research Associate in Computer Vision at the Interdisciplinary Centre for Security, Reliability and Trust (SnT) of the University of Luxembourg. He received his Ph.D. degree in Computer Science from the University of Lille in 2018. His research interests are mainly focused on computer vision and pattern recognition with applications to human behavior understanding.
\end{IEEEbiography}

\vskip -2\baselineskip plus -1fil

\begin{IEEEbiography}[{\includegraphics[width=1in,height=1.20in,clip,keepaspectratio]{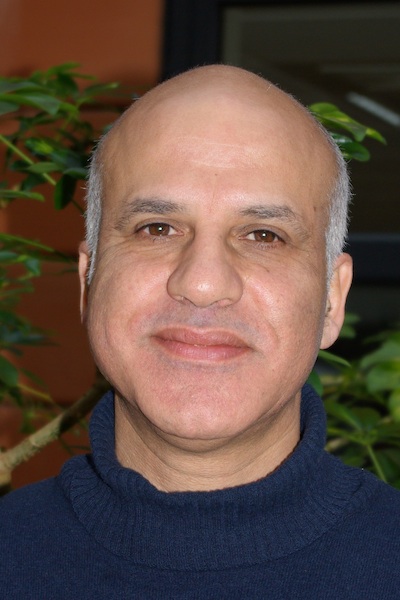}}]{Mohamed Daoudi} is a Full Professor of Computer Science at IMT Lille Douai and CRIStAL (CNRS 9189). He received his Ph.D. degree in Computer Engineering from the University of Lille in 1993. His research interests include pattern recognition and computer vision. He has published over 150 journal and conference articles in these areas. He is AE of IVC Journal, IEEE TMM and Journal of Imaging. He was a General Co-Chair of IEEE FG 2019. He is Fellow of IAPR and IEEE SM.
\end{IEEEbiography}
 
\vskip -2\baselineskip plus -1fil
 
\begin{IEEEbiography}[{\includegraphics[width=1in,height=1.20in,clip,keepaspectratio]{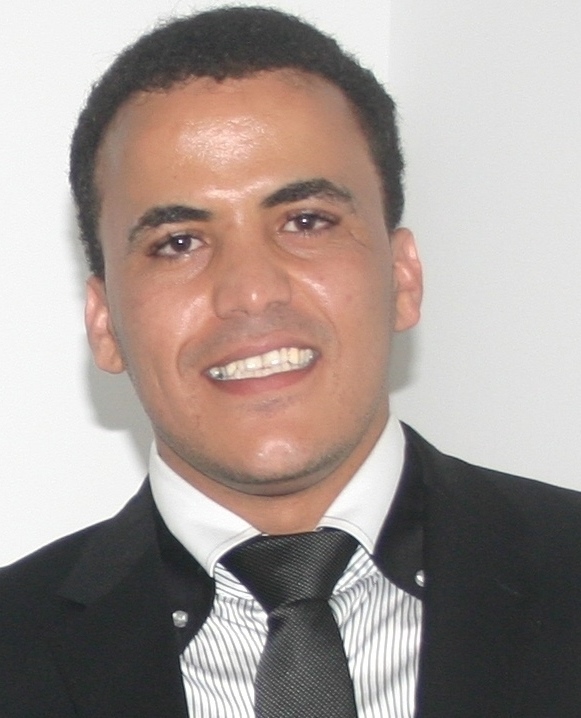}}]{Lahoucine Ballihi}is an Associate Professor at the Department of Computer Science in Faculty of Science, Mohammed V University in Rabat, Morocco. He received his Ph.D degree in Computer Science from the University of Lille, France and the Mohammed V University in Rabat, Morocco in 2012. He is a member of the Computer Science and Telecommunications Research Laboratory of Mohammed V University (LRIT - CNRST URAC 29). His current research interests include computer vision, machine learning, biometrics, image and video analysis and categorization, face and action analysis and recognition, and affective computing. He has published over 25 papers in some of the most distinguished scientific journals and international conferences.
\end{IEEEbiography}

\vskip -2\baselineskip plus -1fil

\begin{IEEEbiography}[{\includegraphics[width=1in,height=1.20in,clip,keepaspectratio]{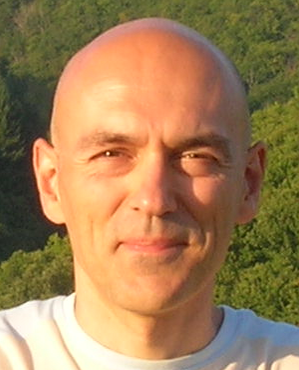}}]{Stefano Berretti} received the Ph.D. in Computer Engineering in 2001. Currently, he is an Associate Professor at University of Florence, Italy. His research interests  are in the areas of pattern recognition, computer vision and multimedia. He has published over 160 conference and journal articles in these areas. He is the Information Director of the ACM Transactions on Multimedia Computing, Communications, and Applications, and Associate Editor of the IET Computer Vision journal.
\end{IEEEbiography}

\newpage
\newpage

\setcounter{section}{0}

\begin{center}
  \textbf{\Large Supplementary Material to the Paper: Automatic Analysis of Facial Expressions Based On Deep Covariance Trajectories}
  
  \vspace{8pt}
  \vspace{8pt}

\end{center}
  \textit{In this supplementary material, we provide the algorithms of our proposed approach, and we present further details on the conducted experiments.}

\section{Algorithms}
For more clarity, we present in this section the algorithms of the proposed approaches. 
For each face $f$, we compute the global and local deep covariance descriptors according to Eq.~(2). Given these descriptors, Algorithm~\ref{algo:CovDescriClassification} summarizes the steps followed to classify the static facial expressions in $Sym^{++}(512)$. 

Concerning the dynamic approach, given a sequence of video frames, we use the same Eq.~(2) to compute the local and global covariance descriptors of each frame, which yields to a global trajectory and four local trajectories for each video. 
For simplicity,  Algorithm~\ref{algo:trajectoriesClasiification} provides a summary of the steps needed to classify the global deep trajectories in $Sym^{++}(m)$, while the same strategy can be extended to classify the local trajectories as in Algorithm~\ref{algo:CovDescriClassification}. 
The equations cited in these algorithms refer to those in the main paper.

\section{Confusion matrices}
\begin{figure}[!ht]
\centering
\begin{minipage}{0.48\linewidth}
\centering
\includegraphics[width=\linewidth]{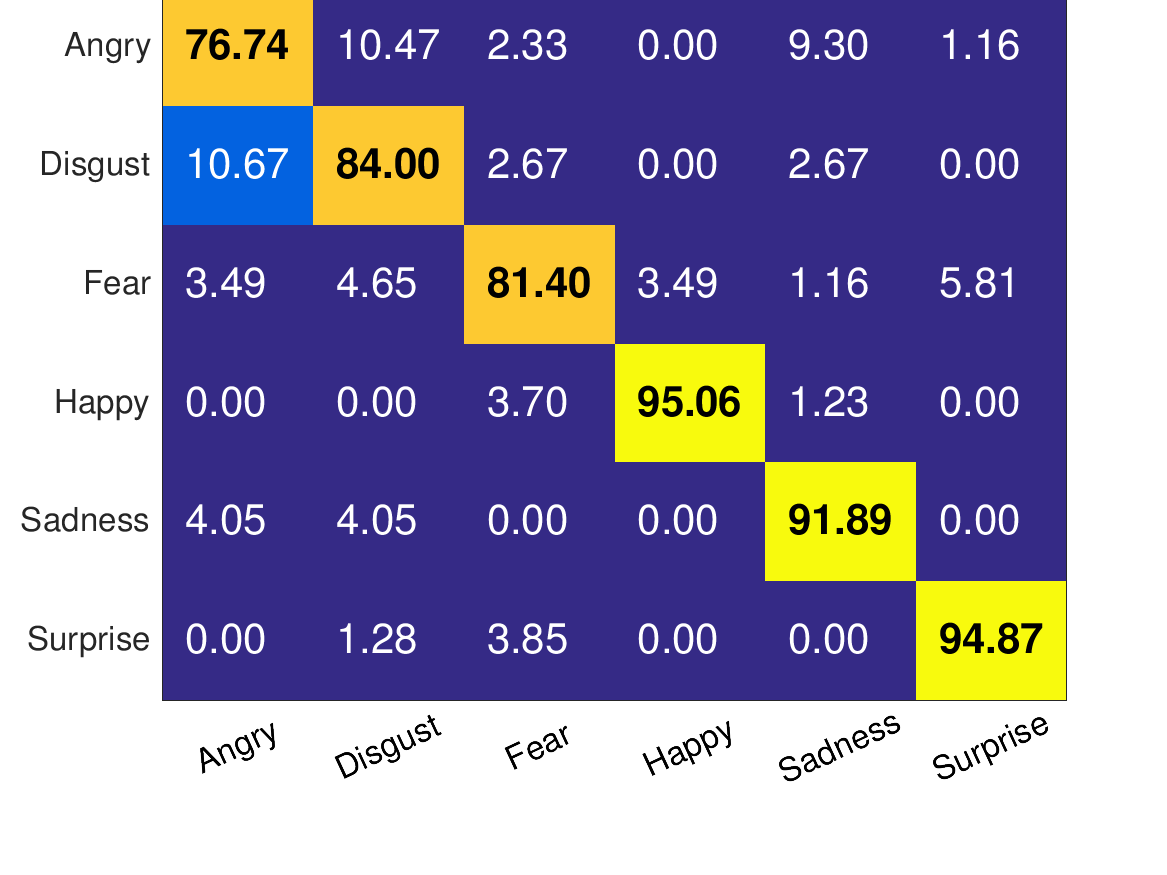}
\end{minipage}
\begin{minipage}{0.48\linewidth}
\centering
\includegraphics[width=\linewidth]{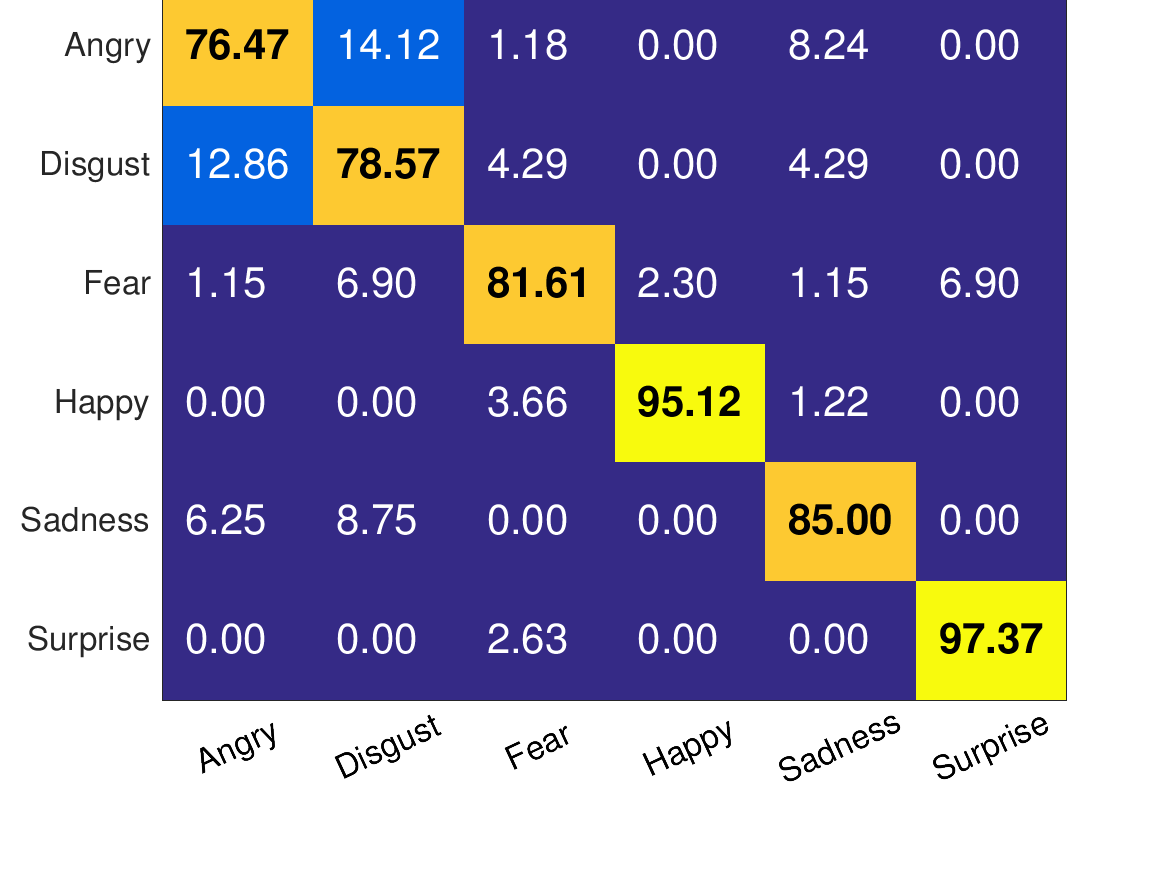}
\end{minipage}
\caption{Confusion matrix on Oulu-CASIA using ExpNet with Kernel fusion. The left panel corresponds to the static approach, while the right one represents the dynamic approach.}
\label{fig:expnet-casia}
\end{figure}

\begin{figure}[!ht]
\centering
\includegraphics[width=0.8\linewidth]{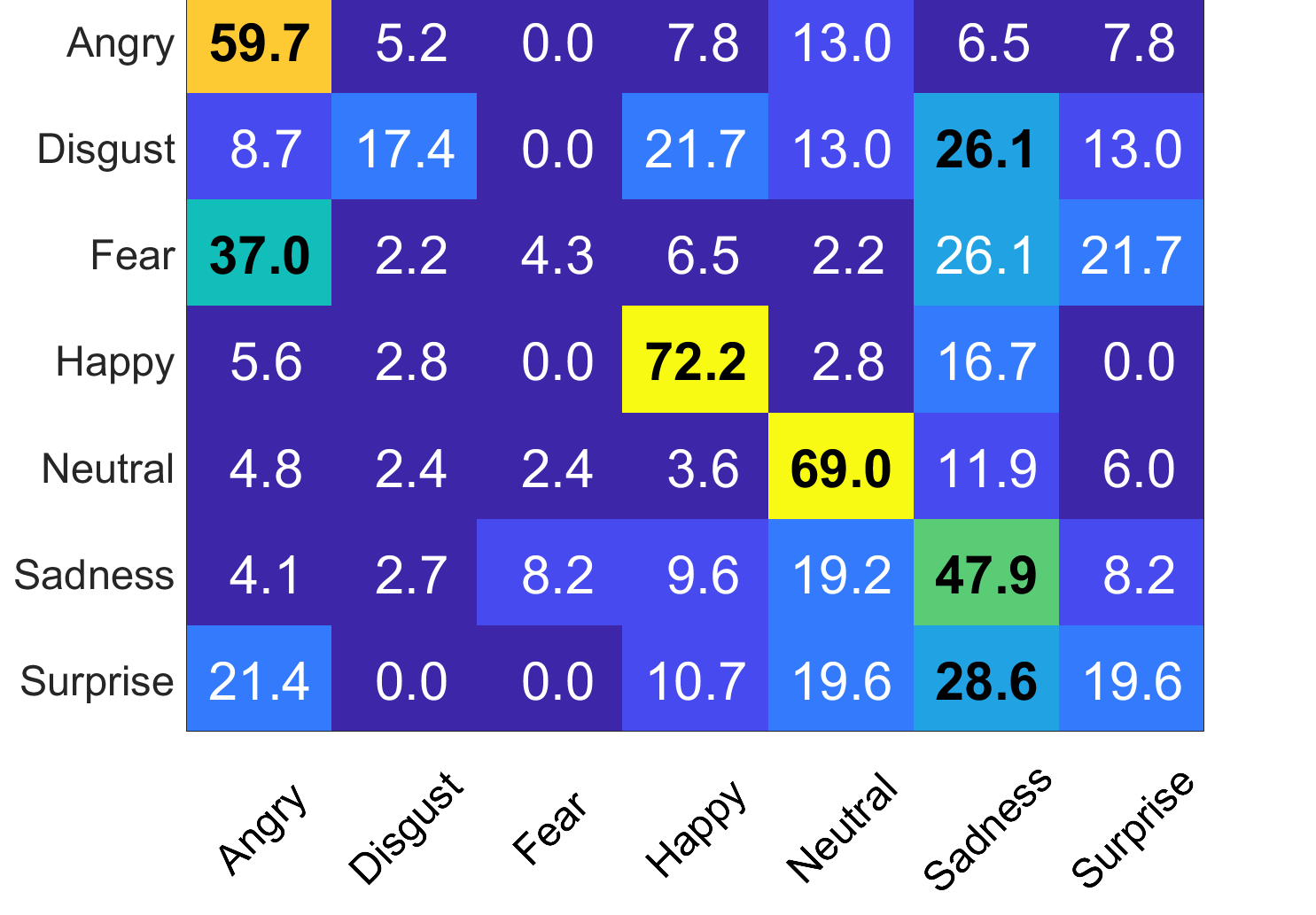}
\caption{Confusion matrix on SFEW for ExpNet with weighted-sum fusion.}
\label{fig:expnet-sfew}
\end{figure}

\begin{figure}[!ht]
\centering
\begin{minipage}{0.48\linewidth}
\centering
\includegraphics[width=\linewidth]{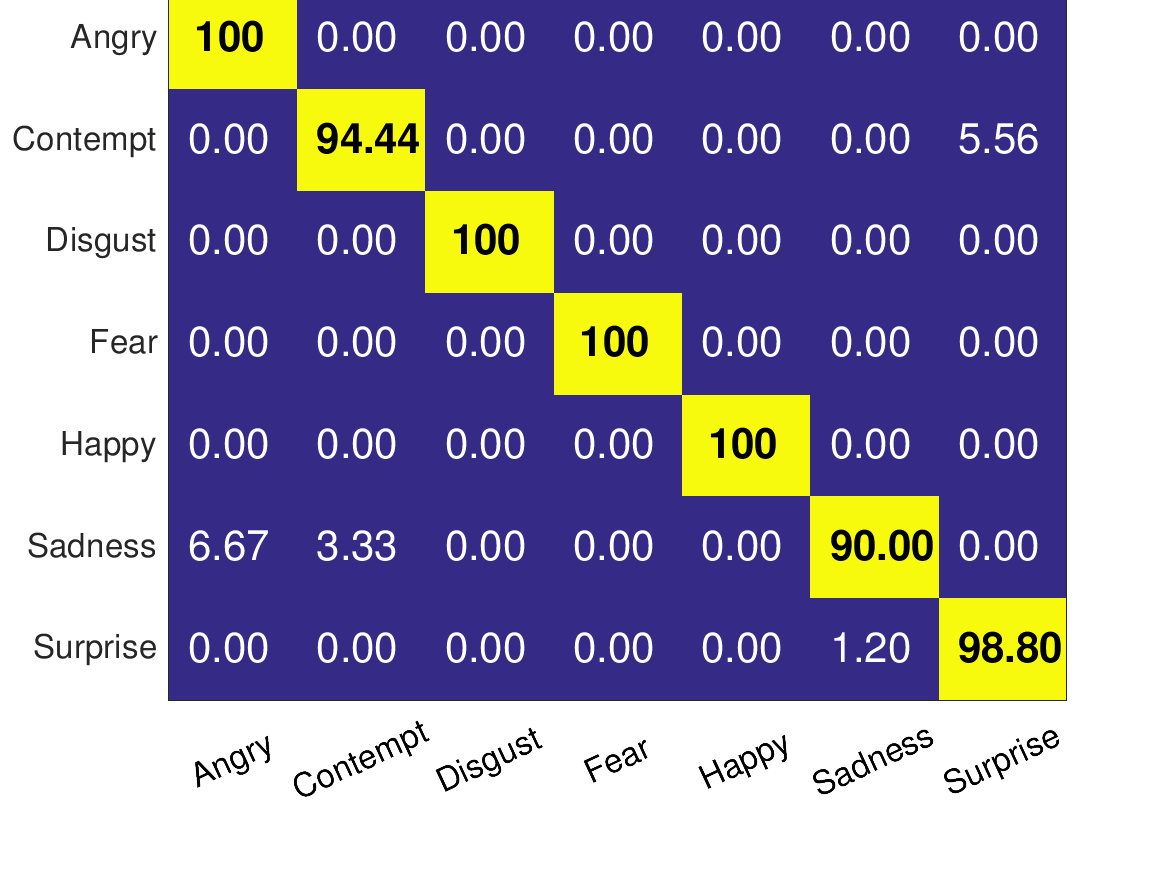}
\end{minipage}
\begin{minipage}{0.48\linewidth}
\centering
\includegraphics[width=\linewidth]{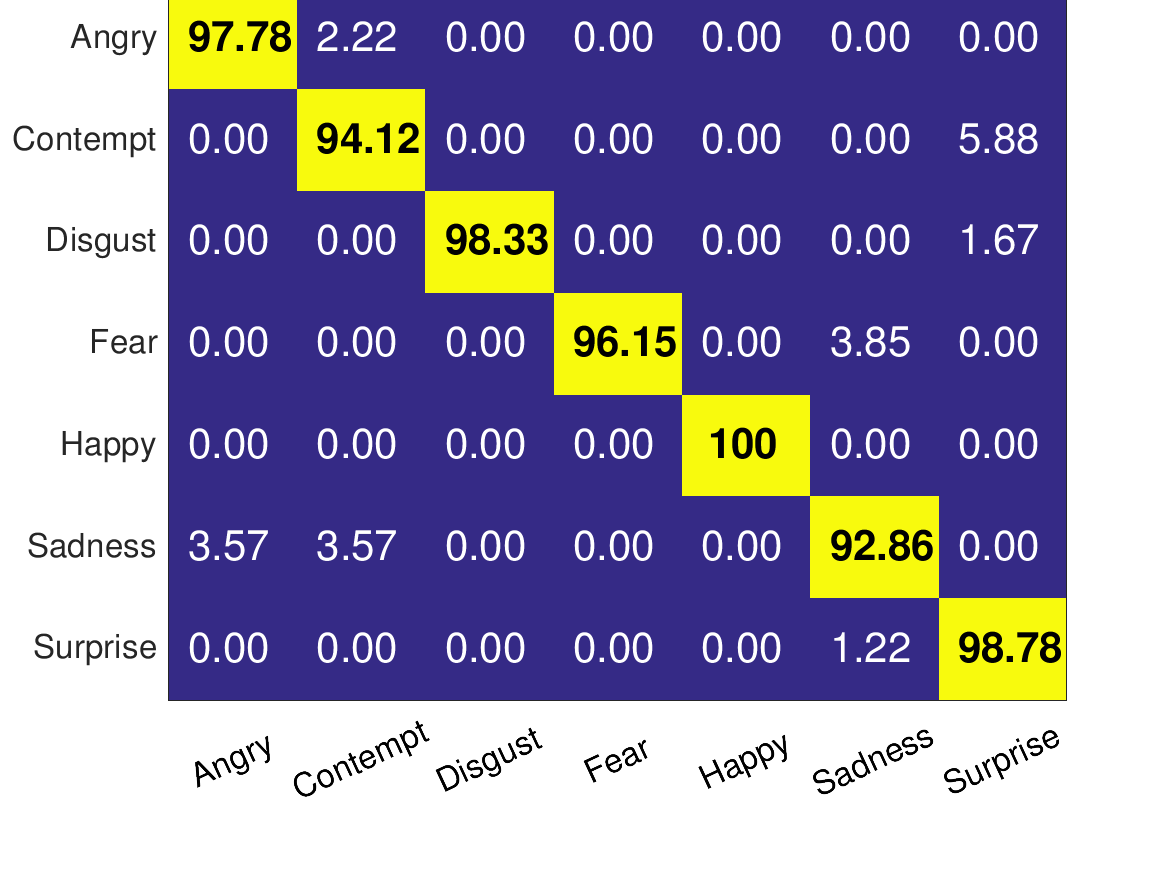}
\end{minipage}
\caption{Confusion matrix on CK+ using ExpNet with Kernel fusion. The left panel corresponds to the static approach, while the right one represents the dynamic approach.}
\label{fig:expnet-ck}
\end{figure}

\begin{figure}[!ht]
\centering
\includegraphics[width=0.8\linewidth]{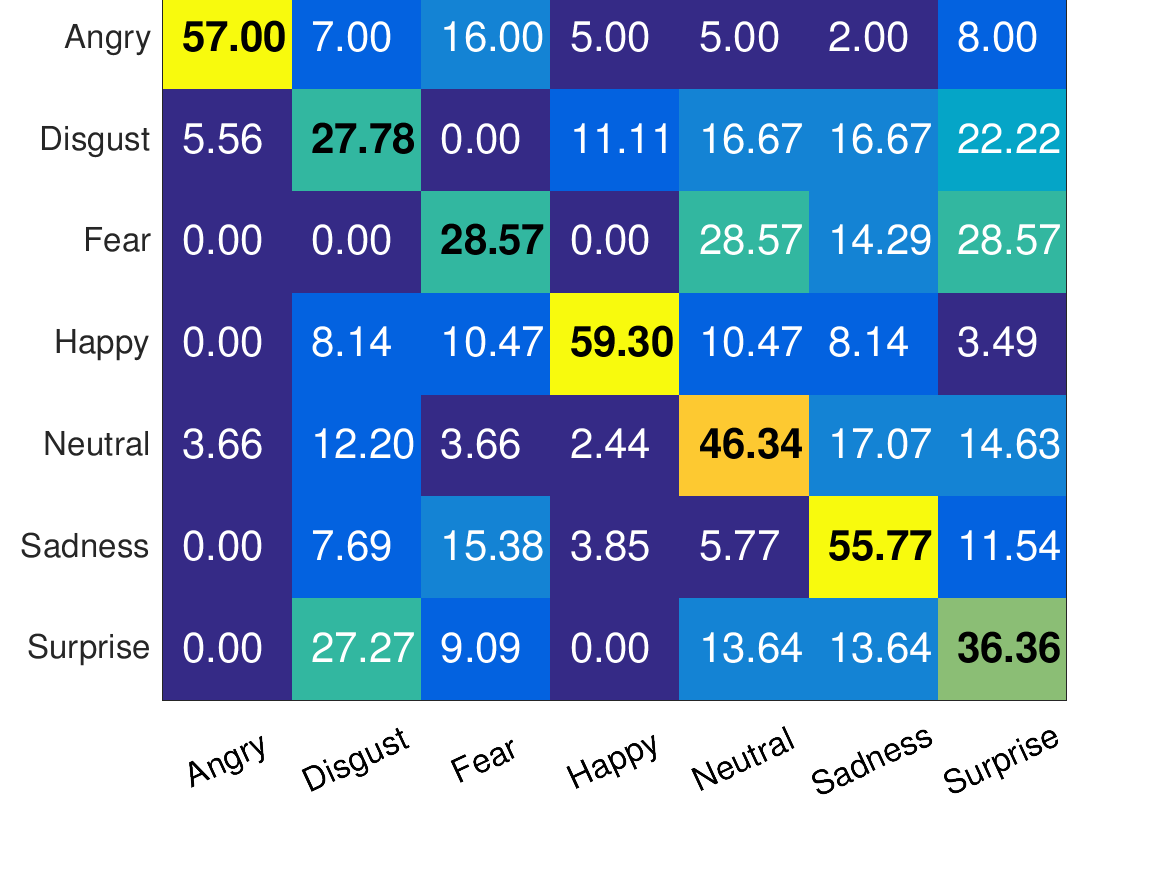}
\caption{Confusion matrix on AFEW for ExpNet with weighted-sum fusion.}
\label{fig:expnet-afew}
\end{figure}

In order to better evaluate our approach, we report in this section the confusion matrices obtained for each dataset used in our experiments. 
The confusion matrices reported here are obtained with the best DCNN model (ExpNet) and our best fusion strategy (Kernel fusion). Figures~\ref{fig:expnet-casia},~\ref{fig:expnet-sfew},~\ref{fig:expnet-ck} and~\ref{fig:expnet-afew} represent the confusion matrices for Oulu-CASIA, SFEW, CK+ and AFEW, respectively.

For Oulu-CASIA, the happy and surprise expressions are better recognized over the rest, while anger and disgust expressions are more challenging. 
The happy expression is the best recognized one also for SFEW and AFEW, followed by the neutral and sad one, while surprise, disgust and fear expressions are harder to recognize. This is encountered in many other works, and it is related to the unbalanced number of expression examples for the different classes included in theses databases as explained in~\cite{ng2015deep}.

Concerning CK+, our approach is able to recognize the majority of the expressions with an accuracy of about $100\%$, except contempt and sadness. 
As for SFEW, this can be explained by the relatively small number of samples for these expressions with respect to the other ones. Table~\ref{tab:numSamples} provides the number of samples representing each facial expression in each dataset. 

\begin{algorithm}[th]
\label{algo:CovDescriClassification}
\SetAlgoLined
 \KwData{$N$ training samples with their associated labels,
 $\{\{C^{R_i}_{\Phi(f_j)}\} _{i=1}^4, y_j\}_{j=1}^N$ and 
 one testing sample $\{C^{R_i}_{\Phi(f)}\} _{i=1}^4$\;}
 \KwResult{Predicted label $y$ of the testing sample}
 \tcc{iterate over four regions}
 \nl \For {$i=1\dots 4$}{
 \tcc{iterate over training examples}
 \nl \For {$j=1\dots N$}{
 \nl \For {$k=1\dots N$}{ 
 \nl Compute $d_{LERM}(C^{R_i}_{\Phi(f_j)},C^{R_i}_{\Phi(f_k)})$ according to Eq.~(3)\; 
 \nl Compute 
 $K^{R_i}(j,k) \leftarrow K(C^{R_i}_{\Phi(f_j)},C^{R_i}_{\Phi(f_k)})$ given by Eq.~(4)\;}
\nl Compute $d_{LERM}(C^{R_i}_{\Phi(f)},C^{R_i}_{\Phi(f_j)})$ according to Eq.~(3)\;
\nl Compute $K_{test}^{R_i}(j) \leftarrow K(C^{R_i}_{\Phi(f)},C^{R_i}_{\Phi(f_j)})$ given by Eq.~(4)\;}
 }
\nl \uIf{Late fusion}{
\nl Train a SVM with each kernel $K^{R_i}$\;
\nl Combine local information using one of Eq.~(5) or Eq.~(6)\;} 

\nl \ElseIf{Early fusion}{\nl Compute kernel $K$ given by one of Eq.~(7) or Eq.~(8)\; 
\nl Train one SVM with the kernel $K$\;}
\nl $y \leftarrow$ SVM with RBF kernel on $Sym^{++}(m)$ using features vectors $\{K_{test}^{R_i} \}_{i=1}^{4}$ fused with the desired fusion strategy\;
\caption{Classification of local covariance descriptors in $Sym^{++}(m)$}
\nl \Return $y$
\end{algorithm}


\begin{algorithm}[th]
\label{algo:trajectoriesClasiification}
\SetAlgoLined
 \KwData{$N_u$ training trajectories $\mathcal{U}=\{ (T^i, Y^i) \}_1^{N_u}$ with their associated labels and one testing trajectory $T_{test}$}
 \KwResult{$Y_{test}$ Predicted label of $T_{test}$}
 \tcc{iterate over training samples}
 \nl \For{$i=1 \dots N_u$}{
 \nl \For{$j=1 \dots N_u$}{
 \nl Align $T^i$ and $T^j$ with Global Alignment\;
 \nl $K(i,j) \leftarrow K_{GA}(T^i, T^j)$ according to Eq.~(13)\;} 
 \nl $K_{test}(i) \leftarrow K_{GA}(T_{test}, T^i)$ according to Eq.~(13)\;}
 \nl Train SVM using kernel $K$\;
 \nl $Y_{test} \leftarrow$ SVM using vector $K_{test}$\;
 \caption{Classification of global deep trajectories in $Sym^{++}(m)$}
\end{algorithm}

\begin{table}[!ht]
\caption{Number of samples for different facial expressions in the Oulu-CASIA, CK+, and SFEW databases}
\scalebox{0.9}{
\begin{tabular}{|c|ccccclcc||c|}
\hline
           & An & Co & Di & Fe & Ha & Ne & Sa & Su & Total \\
           \hline
Oulu-CASIA & 80  & -   & 80  & 80  & 80  & -   & 80  & 80  & 480   \\
CK+        & 45  & 18  & 59  & 25  & 69  & -   & 28  & 83  & 327   \\
SFEW       & 255 & -   & 75  & 124 & 256 & 234 & 150 & 228 & 1322 \\
\hline
\end{tabular}}
\label{tab:numSamples}
\end{table}





\ifCLASSOPTIONcaptionsoff
  \newpage
\fi




\end{document}